\documentclass[10pt,journal,compsoc]{IEEEtran}

\usepackage{enumitem}
\usepackage{xcolor}
\usepackage{graphicx}
\usepackage[caption=false,font=footnotesize,labelfont=sf,textfont=sf]{subfig}
\usepackage[nocompress]{cite}
\usepackage{amsmath,amssymb}
\usepackage[ruled,vlined,linesnumbered]{algorithm2e}
\usepackage{pifont}

\usepackage{adjustbox}
\usepackage{booktabs}

\newcommand{\cmark}{\ding{51}}%
\newcommand{\xmark}{\ding{55}}%

\SetKwInOut{Parameter}{Parameters}
\let\oldnl\nl
\newcommand{\nonl}{\renewcommand{\nl}{\let\nl\oldnl}}

\usepackage{bm}
\usepackage{amsmath, amssymb}

\newcommand{\bmu}{{\bm u}}
\newcommand{\bmv}{{\bm v}}

\newcommand{\bmy}{{\bm y}}

\newcommand{\bmK}{{\bm K}}

\newcommand{\bmS}{{\bm S}}

\newcommand{\bmW}{{\bm W}}

\newtheorem{prop}{Proposition}
\newtheorem{defn}{Definition}

\newtheorem{appr}{Approach}

\newcommand{\transpose}{\textrm{T}}

\newcommand{\R}{\mathbb{R}}

\usepackage{xcolor}

\usepackage{makecell}
\newcommand\w{1.22}
\newcommand{\frameNb}{\makebox[\w cm]{\raggedleft\tiny $-40$} \makebox[\w cm]{\tiny $-30$} \makebox[\w cm]{\tiny $-20$} \makebox[\w cm]{\tiny $-10$} \makebox[\w cm]{\tiny $0$} \makebox[\w cm]{\tiny $+10$} \makebox[\w cm]{\tiny $+20$} \makebox[\w cm]{\tiny $+30$} \makebox[\w cm]{\raggedright\tiny $+40$}}

\newcommand\ws{2.05}
\newcommand{\frameNbs}{\makebox[\ws cm]{\raggedleft $-40$} \makebox[\ws cm]{$-30$} \makebox[\ws cm]{$-20$} \makebox[\ws cm]{$-10$} \makebox[\ws cm]{$0$} \makebox[\ws cm]{$+10$} \makebox[\ws cm]{$+20$} \makebox[\ws cm]{$+30$} \makebox[\ws cm]{\raggedright $+40$}}

\makeatletter
\newcommand{\removelatexerror}{\let\@latex@error\@gobble}
\makeatother

\usepackage{stfloats}
\begin{document}

\title{Diagnosing and Preventing Instabilities in\\ Recurrent Video Processing}

\author{Thomas~Tanay,
        Aivar~Sootla,
        Matteo~Maggioni,
        Puneet~K.~Dokania,
        Philip~Torr,
        Ale\v{s}~Leonardis
        and~Gregory~Slabaugh
\IEEEcompsocitemizethanks{\IEEEcompsocthanksitem T. Tanay, A. Sootla, M. Maggioni and  A. Leonardis are with Huawei Technologies Ltd, Noah’s Ark Lab. E-mail: thomas.tanay@huawei.com
\IEEEcompsocthanksitem G. Slabaugh is with the Queen Mary University of London (work done while at Huawei). P. Torr and P. K. Dokania are with the Department of Engineering Science, University of Oxford.}
}

\IEEEtitleabstractindextext{%
\begin{abstract}
Recurrent models are a popular choice for video enhancement tasks such as video denoising or super-resolution. In this work, we focus on their stability as dynamical systems and show that they tend to fail catastrophically at inference time on long video sequences. To address this issue, we (1)~introduce a diagnostic tool which produces input sequences optimized to trigger instabilities and that can be interpreted as visualizations of temporal receptive fields, and (2)~propose two approaches to enforce the stability of a model during training: constraining the spectral norm or constraining the stable rank of its convolutional layers. We then introduce \emph{Stable Rank Normalization for Convolutional layers} (SRN-C), a new algorithm that enforces these constraints. Our experimental results suggest that SRN-C successfully enforces stability in recurrent video processing models without a significant performance loss.
\end{abstract}

\begin{IEEEkeywords}
Video enhancement, recurrent convolutional neural networks, lipschitz stability, spectral normalization.
\end{IEEEkeywords}
}

\maketitle

\IEEEdisplaynontitleabstractindextext

%
\IEEEpeerreviewmaketitle

\IEEEraisesectionheading{\section{Introduction}\label{sec:introduction}}

\IEEEPARstart{L}{ow-level} computer vision problems such as denoising, demosaicing or super-resolution can be formalised as inverse problems and approached with modern machine learning techniques: a degraded input is processed by a convolutional neural network (CNN) trained in a supervised way to produce a restored output. The input is typically a single frame~\cite{zhang2017beyond,brooks2019unprocessing,gharbi2016deep,kokkinos2019jdd,lim2017enhanced,dai2019second}---but significantly better results can be obtained by leveraging the temporal redundancy of sequential images~\cite{liu2014fast,jo2018deep,wang2019edvr,godard2018deep,frvsr,fuoli2019efficient}. There are two main categories of video processing CNNs. \emph{Feedforward models} operate in a sliding-window fashion and process multiple frames jointly to produce a current output. \emph{Recurrent models} operate in a frame-by-frame fashion but have the ability to store information internally through feedback loops. Recurrent processing is appealing because it reuses information efficiently, potentially over a large number of frames. At the same time, Recurrent Neural Networks (RNNs) are dynamical systems that can exhibit complex and even chaotic behaviors~\cite{laurent2016recurrent}. In the context of sequence modelling for language or sound understanding for instance, RNNs are known to suffer from vanishing and exploding gradient issues at training time~\cite{pascanu2013difficulty}, and to be vulnerable to instabilities through positive feedback at inference time~\cite{miller2019stable}.

\subsection{Motivation}
\label{sec:Motivation}

\begin{figure*}[ht]
\begin{center}
\subfloat{\includegraphics[width=0.93\textwidth]{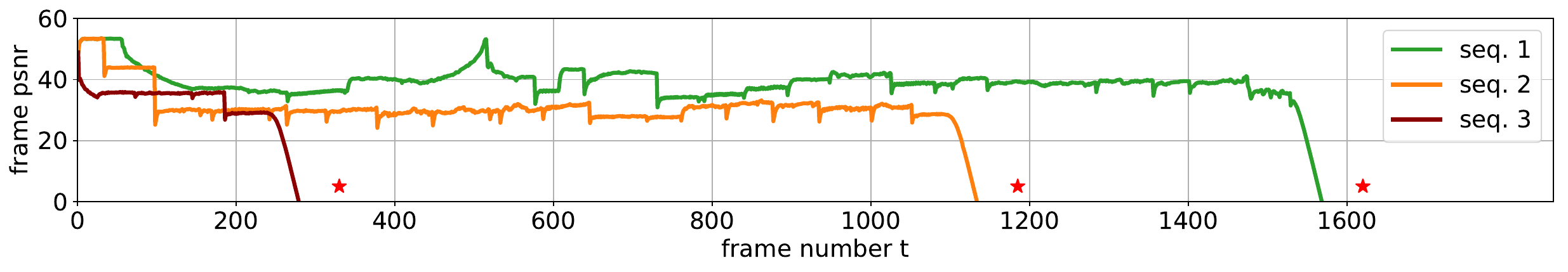}}
\vspace{0.1cm}
\subfloat{\includegraphics[width=0.95\textwidth]{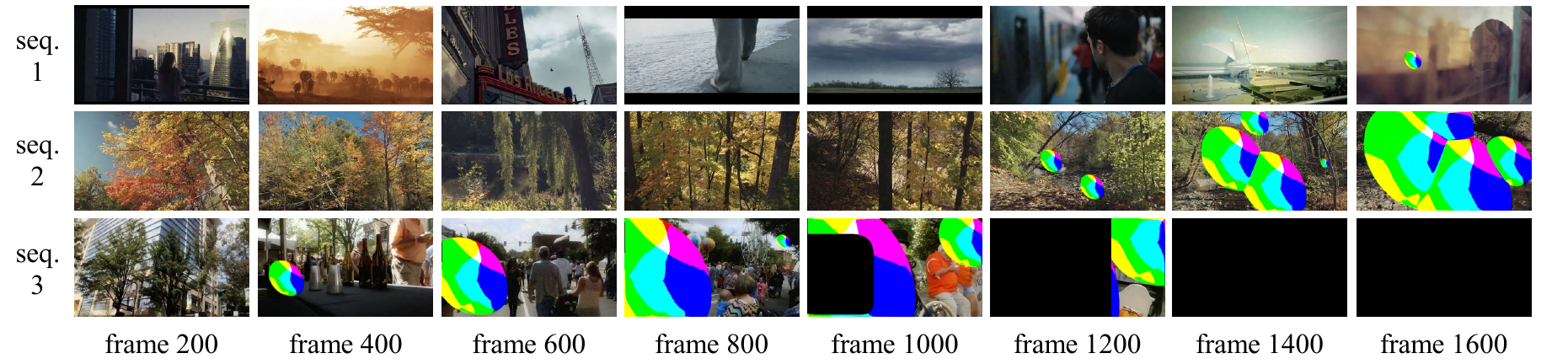}}
\end{center}
\vspace{-0.4cm}
\caption{The recurrent video denoiser from \cite{godard2018deep} is applied to $3$ sequences of $1600$ frames downloaded from vimeo.com. Above: The PSNR per frame is stable for a number of frames varying between $200$ and $1500$, before plunging below $0$ on all $3$ sequences (indicated by red stars). Below: The performance drops manifest themselves in the form of strong colorful artifacts and black masks on the output images (see also Appendix~A).}
\label{instabilities}
\end{figure*}

In the context of video processing too, recurrent CNNs have been observed to suffer from instabilities at inference time on long video sequences. This is the case for instance of the Deep Burst Denoising network of~\cite{godard2018deep} (referred to as DBDNet in the following), consisting of a single-frame branch processing frames independently, and a multi-frame branch where each convolution+ReLU block takes its own output as an additional input at the next time step. To illustrate the instability phenomenon, we retrain DBDNet on the \mbox{Vimeo-90k} dataset~\cite{xue2019video} and we evaluate it on $3$ sequences of $1600$ frames (i.e. roughly one minute of video at $24$ fps) downloaded from vimeo.com. In Figure~\ref{instabilities}, we plot the performance of the model measured by the peak-signal-to-noise-ratio (PSNR) as a function of the frame number and we observe instabilities on all $3$ sequences: the PSNR plunges permanently at some unpredictable time in the sequence. Visually, these instabilities correspond to the formation of colorful artifacts at random locations, growing locally until the entire output frame is covered. Numerically, they correspond to diverging or saturating pixel values.

\begin{figure*}[ht]
\begin{center}
\subfloat[From left to right: feature-shifting~\cite{huang2015bidirectional,lin2019tsm}, feature-recurrence~\cite{huang2015bidirectional,chen2016deep,godard2018deep}, frame-recurrence~\cite{frvsr,arias2019kalman}, RLSP~\cite{fuoli2019efficient}.\label{random_models_architectures}]{%
\includegraphics[width=0.47\textwidth]{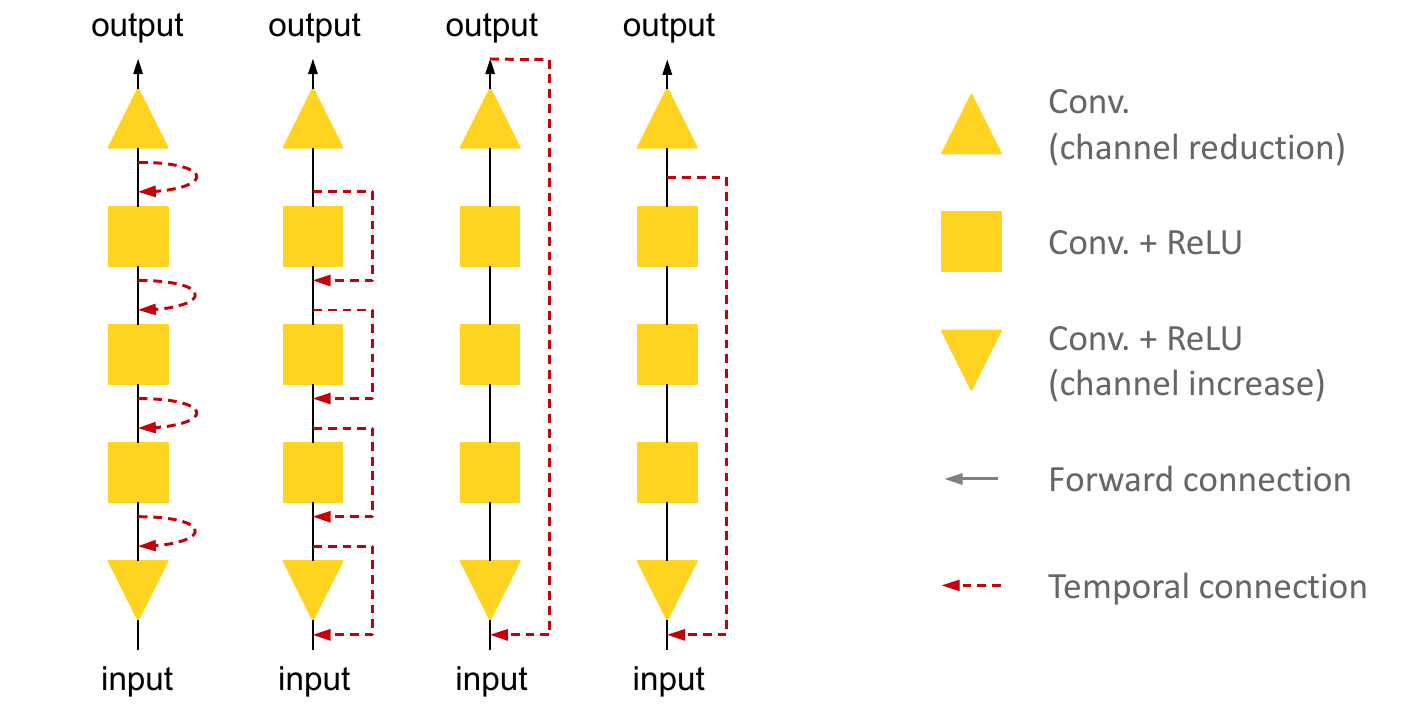}
}
\hfill
\subfloat[Norm of the output frame over a random sequence of 50 frames, for six architectures initialized with a Gaussian distribution $(\sigma=0.1)$.\label{random_models_plots}]{%
\includegraphics[width=0.47\textwidth, height=0.235\textwidth]{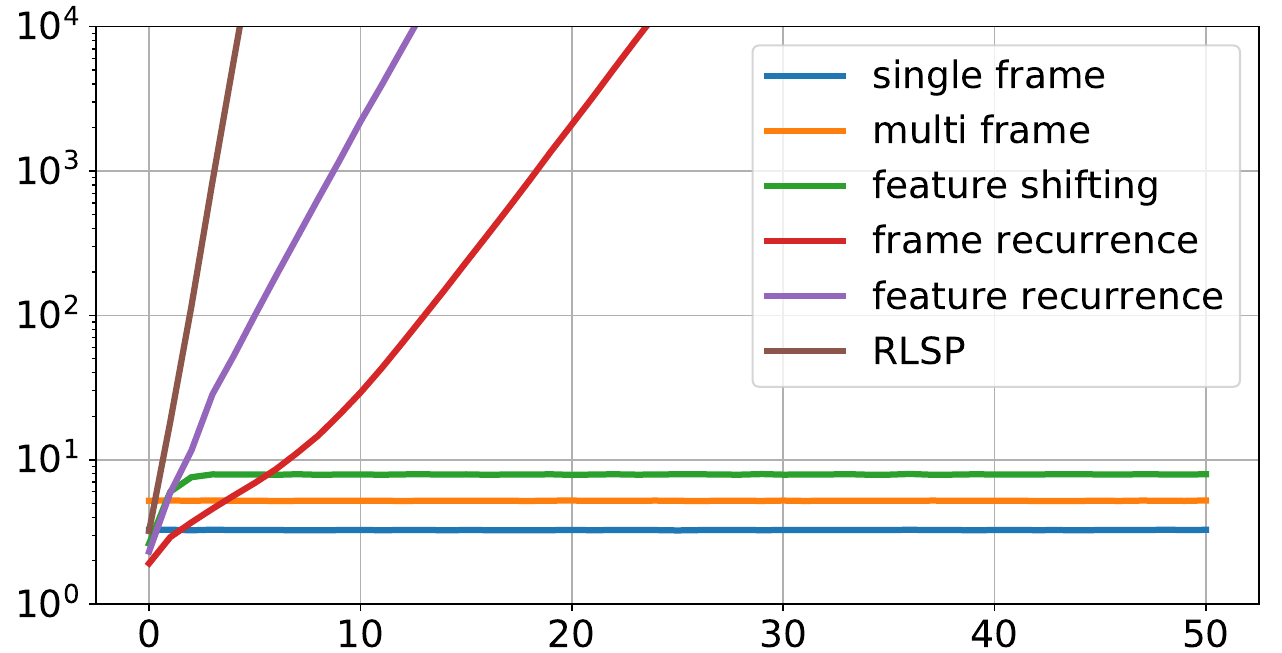}
}
\end{center}
\vspace{-0.2cm}
\caption{For untrained models over random inputs, feedforward architectures produce stable outputs (single-frame, multi-frame, feature-shifting) while recurrent architectures diverge (frame-recurrence, feature-recurrence, RLSP).}
\label{random_models}
\end{figure*}

Our working hypothesis is that recurrent connections create positive feedback loops prone to this type of divergent behaviour. As a proof of concept, we consider a backbone architecture made of five convolutional layers interleaved with ReLU non-linearities, and we augment it with various strategies for temporal processing. We consider single-frame and multi-frame inputs, and four types of temporal connections inspired from existing video processing works: \emph{feature-shifting} where features are extracted and fed back at the same level~\cite{huang2015bidirectional,lin2019tsm}, \emph{feature-recurrence} where features are extracted and fed back at a lower level~\cite{huang2015bidirectional,chen2016deep,godard2018deep}, \emph{frame-recurrence} where the output frame is fed back as an input~\cite{frvsr,arias2019kalman} and \emph{recurrent latent space propagation} (RLSP) where the latent space at a high level is fed back as an input~\cite{fuoli2019efficient} (see Figure~\ref{random_models_architectures}). We then initialize all the models randomly and feed them with random inputs. We see that feedforward architectures (single-frame, multi-frame, feature-shifting) produce stable outputs while recurrent architectures (feature-recurrence, frame-recurrence, RLSP) produce outputs that diverge (see Figure~\ref{random_models_plots}). Note that feature-shifting is non-recurrent since information cannot flow indefinitely inside a feedback loop.

The instability phenomenon described here is a serious concern for the deployment of recurrent video processing models in real-world applications. A number of coping strategies can be considered to operate an unstable model in a stable manner at inference time but none of them are truly satisfying (see Section~\ref{sec: Discussion}). In this paper, we propose to solve the instability problem altogether by enforcing mathematically derived stability constraints during training.

\subsection{Contributions}

The main contributions of this paper are as follows:
\begin{itemize}[leftmargin=*]
    \item We identify a serious vulnerability affecting RNNs for video processing: they can be unstable at inference time and fail catastrophically on long video sequences.
    \item To test stability, we introduce a fast and reliable diagnostic tool that produces input sequences optimized to trigger instabilities, and that can be interpreted as temporal receptive fields.
    \item We investigate two approaches to enforce the stability of recurrent video processing networks: constraining the spectral norm or constraining the stable rank of their convolutional layers. 
    \item We extend a recently proposed weight normalization scheme called Stable Rank Normalization (SRN)~\cite{sanyal2020stable} that simultaneously constrains the spectral norm and the stable rank of any linear mapping, to convolutional layers. We call it Stable Rank Normalization for \emph{Convolutional layers} (SRN-C)---as opposed to stable rank normalization applied to the \emph{convolutional kernel}.
\end{itemize}

\subsection{Related Work}

A number of approaches have been proposed in the literature for extending the connectivity of a CNN in the time domain. In~\cite{karpathy2014large}, the authors identify three classes of feedforward models: \emph{early fusion} where the frames over a fixed time window are concatenated and processed simultaneously, \emph{late fusion} where the frames are processed independently and only their latent space representations are concatenated, and \emph{slow fusion} where intermediary features are concatenated at multiple levels such that higher layers get access to progressively more global information in both spatial and temporal dimensions. Variants of slow fusion were introduced a number times under different names: \emph{conditional convolutions} in~\cite{huang2015bidirectional}, \emph{3D~convolutions} in~\cite{tran2015learning}, \emph{progressive fusion} in~\cite{yi2019progressive} and \emph{feature-shifting} in~\cite{lin2019tsm} all fuse features from different time steps at multiple levels of the network. For video restoration tasks, most standard models implement a form of early fusion~\cite{liu2014fast,caballero2017real,jo2018deep,mildenhall2018burst,wang2019edvr} but late fusion~\cite{tassano2019dvdnet} and two-level fusion~\cite{tassano2020fastdvdnet} have also been used.

In contrast with the feedforward fusion approaches above, recurrent models contain feedback loops where the features are fed back to the same processing block multiple times. One of the earliest applications of RNNs in video restoration was in~\cite{huang2015bidirectional}. The architecture proposed for video super-resolution used a large number of temporal connections, with forward and backward subnetworks processing inputs in temporal and reverse-temporal order, each using both \emph{conditional} and \emph{recurrent} connections corresponding to \emph{feature-shifting} and \emph{feature-recurrence} respectively according to our taxonomy from Figure~\ref{random_models_architectures}. Feature recurrence was used again for video denoising in~\cite{chen2016deep} on a deep but non-convolutional RNN, and in~\cite{godard2018deep} on the multi-frame branch of a hybrid architecture constituted of a single-frame denoiser and a multi-frame denoiser. Frame-recurrence, where the previous output frame is fed back as an additional input at the next time step, was introduced for video super-resolution in~\cite{frvsr}. This type of recurrence was studied further in~\cite{arias2019kalman} where a connection was made with the concept of Kalman filtering~\cite{kalman1960new}. Recently and still in super-resolution, recurrent latent space propagation (RLSP) was introduced~\cite{fuoli2019efficient}. RLSP can be interpreted as maximizing the depth and width of the recurrent connection, compared to feature-recurrence and frame-recurrence. Iterative approaches~\cite{zamir2017feedback,haris2018deep,haris2019recurrent} are conceptually similar to recurrent ones, but the feedback loop is part of a refinement mechanism that occurs for a fixed number of iterations, chosen as a hyperparameter by the user independently of the temporal length of the video sequence. State-of-the-art performance in video restoration has regularly shifted between feedforward and recurrent architectures in the literature~\cite{godard2018deep,wang2019edvr,frvsr,tassano2020fastdvdnet,yue2020supervised,chan2021basicvsr}, the current state-of-the-art~\cite{Yang2021NTIRE2C,Son2021NTIRE2C} making use of multiple recurrent connections~\cite{chan2021basicvsr,chan2021basicvsr++}. We illustrate the advantage of using a recurrent architecture over a feedforward one in Appendix~B.

Training Recurrent Neural Networks (RNNs) is notoriously difficult due to the \emph{vanishing and exploding gradients} problem: RNNs are trained by unrolling through time, which is effectively equivalent to training a very deep network~\cite{bengio1994learning,pascanu2013difficulty}. Relatedly, \emph{RNNs are vulnerable to instabilities at inference time on long sequences}. This phenomenon was studied in the context of 1-layer fully connected networks in~\cite{jin1994absolute}, and in the context of multi-layer and LSTM networks in~\cite{miller2019stable}, where it was shown that the RNN is stable if its Lipschitz constant is less than~$1$. In~\cite{miller2019stable}, it was proposed to enforce this stability constraint by projecting onto the spectral norm ball of the recurrence matrix (i.e. by clipping its singular values to 1) and a number of recent works have sought to avoid vanishing and exploding gradients by enforcing orthogonality (i.e. setting all the singular values to 1)~\cite{arjovsky2016unitary,wisdom2016full,mhammedi2017efficient,vorontsov2017orthogonality,jose2018kronecker,zhang2018stabilizing}. In the context of CNNs however, enforcing the Lipschitz constraint is challenging. In~\cite{sedghi2019singular}, it was proposed to clip singular values of the convolutional (but non-recurrent) layer, which was flattened into a matrix using the doubly block circulant matrix representation. However, the optimization method does not have formal convergence guarantees and requires computing all singular values of the flattened kernel. In~\cite{miyato2018spectral}, it was proposed to normalize the kernel of the convolutional (but non-recurrent) layer during training, the $4$-D kernel being first reshaped into a \mbox{$2$-D} matrix by flattening its first three dimensions. Normalization is performed by an elegant iterative scheme employing the power iteration estimating the maximal singular value of the flattened kernel. However, as discussed in~\cite{scaman2018lipschitz} and~\cite{gouk2018regularisation}, this approach is not suitable for Lipschitz regularization due to the invalid flattening operation used, and as a result is not suitable for stability enforcement using~\cite{miller2019stable} either. To solve this issue, Gouk et al.~\cite{gouk2018regularisation} suggested replacing the \mbox{$2$-D} matrix products in the power iteration with convolution and transpose convolution operations using the \mbox{$4$-D} kernel tensor directly. This method was applied with success in \cite{behrmann2019invertible} to train invertible ResNets. 

Recently, Sanyal et al. (2020)~\cite{sanyal2020stable} proposed Stable Rank Normalization (SRN), a provably optimal weight normalization scheme which minimizes the stable rank of a linear operator while constraining the spectral norm. They showed that SRN, while improving the classification accuracy, also improves generalization of neural networks and reduces memorization. However, SRN operates on a 2-D reshaping of the convolutional kernel, instead of operating on the convolutional layer as a whole.

\section{Stability in recurrent video processing}

In this section we define the notion stability, we introduce the Temporal Receptive Field (TRF) diagnostic tool and the two stability constraints, and we present our Stable Rank Normalization for Convolutional layers algorithm (SRN-C).

\subsection{Definitions}
\label{sec:Definitions}

Partially reusing notations from~\cite{miller2019stable}, we define a recurrent video processing model as a non-linear dynamical system given by a Lipschitz continuous \emph{recurrence map} \mbox{$\phi_w : \mathbb{R}^n \times \mathbb{R}^d \to \mathbb{R}^n$} and an \emph{output map} $\psi_w : \mathbb{R}^n \to \mathbb{R}^d$ parameterized by $w \in \mathbb{R}^m$. The hidden state $h_t \in \mathbb{R}^n$ and the output image $y_t \in \mathbb{R}^d$ evolve in discrete time steps according to the update rule\footnote{The case where $y_t = h_t$ corresponds to the frame-recurrent architecture of~\cite{frvsr}.}
\begin{equation}
\begin{cases}
  h_t &= \phi_w(h_{t-1}, x_t)\\
  y_t &= \psi_w(h_t)
\end{cases}
\end{equation}
where the vector $x_t \in [0,1]^d$ is an arbitrary input image provided to the system at time~$t$. 

In Section~\ref{sec:Motivation}, we showed examples of models that produced diverging outputs and called them ``unstable''. In the following, we propose to use the notion of \emph{Bounded-Input Bounded-Output (BIBO) stability} to formalize this behaviour. 
\begin{defn}\label{def:stability}
A recurrent video processing model is \emph{BIBO stable} if, for any admissible input $\{x_t\}_{t= 0}^\infty$ for which there exist a constant $C_1$ such that $\sup_{t\ge 0}\|x_t\|\le C_1$, there exists a constant $C_2$ such that $\sup_{t\ge 0}\|y_t\|\le C_2$.
\end{defn}
This definition is well suited for models using ReLU activation functions and the diagnostic tool we introduce in the next section relies on it. However, it fails to capture a stricter notion of stability for models with bounded activation functions, which are BIBO stable by construction\footnote{Simply applying a sigmoid function to the output of an unstable model technically makes it BIBO stable, yet in practice, the model still suffers from instabilities and its output simply saturates.}. Therefore, we will use the stricter notion of \emph{Lipschitz stability} for stability enforcement, as in~\cite{miller2019stable}.
\begin{defn}\label{def:contractiveness}
A recurrent video processing model is \emph{Lipschitz stable} if its recurrence map $\phi_w$ is \emph{contractive in $h$}, i.e. if there exists a constant $L<1$ such that, for any states $h, h^\prime\in \R^n$ and input $x \in \R^d$, 
\begin{equation}
    \| \phi_w(h, x) - \phi_w(h^\prime, x)\| \leq L \| h - h^\prime \|.
\end{equation}
\end{defn}
The constant $L$ is called the Lipschitz constant of $\phi_w$. We show easily that Lipschitz stability implies BIBO stability, but the reciprocal is not always true. 

\subsection{Diagnosis}
\label{sec:Diagnosis}

Consider a \emph{trained} recurrent video processing model $(\phi_w, \psi_w)$. A prerequisite to use it in real-world applications is to determine whether it is stable or not. Unfortunately, proving that a model is \emph{BIBO stable} is difficult: in principle, this requires to perform an exhaustive search over the (infinite) set of valid inputs, and check that none of them are unstable. Alternatively, one could try to show that the model is \emph{Lipschitz stable} instead. However, computing the Lipschitz constant for a neural network is, in general, NP-hard~\cite{scaman2018lipschitz} and therefore intractable. 

In practice, one realistic test is to run the model on hours of video data and report possible instabilities---effectively performing a \emph{random search} for unstable sequences over the set of valid inputs. When an unstable sequence is found, this test constitutes a formal guarantee that the model is unstable. When no unstable sequence is found, however, nothing can be concluded with certainty: the model could be stable, or the search could simply have failed. It is not clear what type of input data should be used and how long the search should last before concluding reliably that the model is, indeed, stable. As Figure~\ref{instabilities} shows, these are not trivial questions: instabilities do not occur after the same number of frames on all video sequences and it can easily take more than a thousand frames before an instability occurs.

\begin{figure*}
\centering
\subfloat[TRF of an unstable model. The receptive field is not temporally bounded and the output sequence $Y$ diverges. \label{unstable-TRF}]{%
\thead{\vspace{-0.4cm}\\ $t$\vspace{0.15cm}\\ $X$\vspace{0.2cm}\\ $Y$} \hspace{-0.6 cm} \thead{\frameNbs\\ \includegraphics[width=0.95\textwidth, height=1cm]{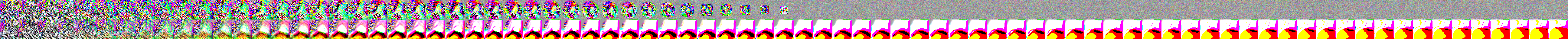}}}

\subfloat[TRF of a stable model. The receptive field is temporally bounded ($\approx 17$ frames) and the output sequence $Y$ is well-behaved. \label{stable-TRF}]{%
\thead{\vspace{-0.4cm}\\ $t$\vspace{0.15cm}\\ $X$\vspace{0.2cm}\\ $Y$} \hspace{-0.6 cm} \thead{\frameNbs\\ \includegraphics[width=0.95\textwidth, height=1cm]{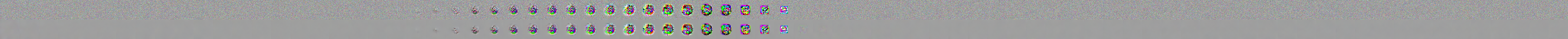}}}
\caption{Temporal Receptive Field (TRF) as a diagnostic tool. The input sequence $X$ is optimized to trigger instabilities in the output sequence~$Y$. The sequences have been horizontally compressed to fit the page width. In the rest of the paper, we plot TRFs every 5 frames for convenience, but the optimization is always performed on sequences of 81 frames ($\tau = 40$).}
\end{figure*}

Here, we propose to approach the problem in a different way and to \emph{search for unstable sequences by gradient descent}. We introduce a stress test that actively tries to trigger instabilities by maximising the output of the RNN at a given time step with respect to its temporally unrolled input. More precisely, we fix a sequence length $2\tau + 1$ and an image size $d$, and consider the finite input sequence $X = (x_{-\tau}, ..., x_{\tau})$ with the corresponding finite output sequence $Y = (y_{-\tau}, ..., y_{\tau})$ such that $h_{-\tau-1} = 0$ (i.e. the initial hidden state is null) and for all $t \in [-\tau,\tau]$:
\begin{alignat*}{2}
&h_t = \phi_w(x_t, h_{t-1}) & \quad\quad  & \text{$\phi_w$ is unrolled over the sequence}\\
&y_t = \psi_w(h_t) &      & \text{$\psi_w$ maps to the output image}
\end{alignat*}
We then search for an unstable sequence by optimizing: 
\begin{equation}\label{adv-sequence}
\max_{0 \leq X \leq 1} \;\; |p|
\end{equation}
where $p$ is the pixel in the centre of $y_0$, the output frame at time $t=0$. In words, we search for an input sequence $X$ such that the corresponding output sequence $Y$ diverges maximally in $p$. This optimization process affects all the pixels in $X$ having an influence on $p$, revealing the flow of information from past pixels to the current one, and therefore it can be interpreted as a visualization of the \emph{Temporal Receptive Field} (TRF) of the model. Computing the TRF can then be used as a diagnostic tool for stability, and we observe two possible behaviours. 
\begin{itemize}[leftmargin=*,topsep=4pt, itemsep=2pt]
\item \textbf{The TRF is not temporally bounded.} Input frames in the distant past have an effect on $p$ and output frames in the distant future diverge (see Figure~\ref{unstable-TRF}). The input sequence $X$ constitutes an unstable sequence and we can conclude with certainty that the model is unstable.
\item \textbf{The TRF is temporally bounded.} Input frames in the distant past have no effect on $p$ and output frames in the distant future remain unaffected (see Figure~\ref{stable-TRF}). No unstable sequence has been found and we can conclude with reasonable confidence that the model is stable.
\end{itemize}
This type of optimization on the output of a model with respect to its input is related to the work on \emph{adversarial examples} in image classification~\cite{szegedy2014intriguing,goodfellow2015explaining,kurakin2017adversarial,ilyas2019adversarial,biggio2018wild} 
and on \emph{feature visualization}~\cite{erhan2009visualizing,szegedy2014intriguing,mahendran2015understanding,nguyen2015deep,olah2017feature}. To the best of our knowledge however, it has never been used in the context of recurrent networks and the use we make of it here to test the temporal stability of a model is novel. In our experiments, we initialise $X$ randomly, we choose a sequence length $2\tau + 1 = 81$ and an image size $d = 64 \times 64$. We then solve the optimization problem using the Adam optimizer for 500 iterations. This test typically takes a few minutes to complete, which is much faster and more computationally efficient than running the model on hours of video data (2h of video data takes approximately 1h to process at 50 fps). For this reason, it is particularly adapted to perform model invalidation quickly.

As discussed before, neither running the model on hours of video data (random search) nor computing the TRF (gradient descent search) can guarantee stability with certainty, but we show in the experimental section that the two tests give consistent answers on the stability of various models---providing positive evidence that they are able to identify stable models correctly. TRFs also help visualize the temporal window of influence of a model, or how long information can stay in memory, and therefore illustrates the relationship between stability and memory in RNNs.

\subsection{Prevention}
\label{sec:Prevention}

Now, consider an \emph{untrained} recurrent video processing model $(\phi_w,\psi_w)$. In order to prevent instabilities from occurring at inference time, we want to enforce a stability constraint into the model during training. As discussed in Section~\ref{sec:Definitions}, this can be achieved by ensuring that $\phi_w$ is contractive with respect to the recurrent variable.

Suppose that $\phi_w$ is made of $l$ convolutional layers separated by ReLU non-linearities. Each convolution can be represented by its 4D kernel tensors $\bmK$, or by a corresponding 2D matrix $\bmW$ obtained from $\bmK$ as a block matrix of doubly-block-circulant matrices~\cite{sedghi2019singular}. Then for a layer $\bmW$ with singular values $\{\sigma_k\}$ assumed to be sorted, the spectral norm is $\|\bmW\| = \sigma_1$, the Frobenius norm is $\|\bmW\|_F = \sqrt{\sum_k \sigma_k^2}$ and the \emph{stable rank}\footnote{The stable rank is a soft, numerical approximation of the rank operator. It is \emph{stable} under small perturbations of the matrix---the name has nothing to do \emph{a priori} with the notion of stability studied here.} is defined as~\cite{rudelson2007sampling,sanyal2020stable}:
\begin{equation}
    \text{srank}(\bmW) = \frac{\|\bmW\|_F^2}{\|\bmW\|^2} = \frac{\sum_k \sigma_k^2}{\sigma_1^2}.
\end{equation}
It is a scale independent quantity and can be interpreted as an \emph{area-under-the-curve} for the normalized singular value spectrum. Now, let $L$ be the Lipschitz constant of $\phi_w$. Since the Lipschitz constant of the ReLU non-linearity is~1, we know that $L$ is upper-bounded by the product of the spectral norms of the linear layers~\cite{scaman2018lipschitz,szegedy2014intriguing}.
\begin{prop}
For a recurrent model $\phi_w$ constituted of $l$ linear layers with weight matrices $\bmW_1, ..., \bmW_l \in \mathbb{R}^{n\times n}$ interspaced with ReLU non-linearities, the Lipschitz constant $L$ of $\phi_w$ satisfies:
\begin{equation}\label{eq:Lipschitz-upper-bound}
    L \leq \prod_{i=1}^l \|\bmW_i\|.
\end{equation}
\end{prop}
Using this upper-bound, one can guarantee that $\phi_w$ is contractive (i.e. $L < 1$) with the following approach.
\begin{appr}
\textbf{Hard Lipschitz Constraint}\nopagebreak

\noindent For all $i \in [1,l]$, we enforce $\|\bmW_i\| < 1$.
\end{appr}
This approach has the advantage of providing a theoretical guarantee of stability. However, it is overly restrictive because the upper-bound~(\ref{eq:Lipschitz-upper-bound}) tends to significantly overestimate the Lipschitz constant $L$~\cite{sanyal2020stable}. To illustrate why this is the case, suppose that $\phi_w$ contains two layers $\bmW_1$ and $\bmW_2$. Then the only situation in which we have $L = \|\bmW_1\| \|\bmW_2\|$ is when the first right singular vector of $\bmW_1$ is aligned with the first left singular vector of $\bmW_2$. In other situations, $L$ depends on the rest of the singular value spectra of $\bmW_1$ and $\bmW_2$ and hence, on their stable ranks. These considerations lead us to a second approach to enforce $L < 1$.
\begin{appr}
\textbf{Soft Lipschitz Constraint}\nopagebreak

\noindent For all $i \in [1,l]$, we fix $\|\bmW_i\| = \alpha$ and minimize $\text{\normalfont srank}(\bmW_i)$.
\end{appr}
This approach does not offer any theoretical guarantee of stability for $\alpha > 1$. However, we verify empirically in Section~\ref{sec:Experiments} that it is also successful at promoting stability.

\begin{figure*}
\noindent
\begin{minipage}[t]{0.49\textwidth}
\removelatexerror
\footnotesize
\begin{algorithm}[H]
\DontPrintSemicolon
\KwIn{Number of iterations $N$, learning rate $\eta$,\;
\hspace{1cm} number of channels $m$, image size $n$,\;
\hspace{1cm} initial $\bmK \in \mathbb{R}^{k\times k\times m \times m}$, initial $\bmu \in \mathbb{R}^m$.}
\Parameter{Spectral norm $\alpha$, stable rank $\beta$.}
\nonl \Begin{
    \nonl \For{$i=1,\dots, N$}{
    $\widetilde{\bmK} = \text{Reshape}(\bmK,[k k m,m])^\transpose$\;
    \nonl \emph{Power iteration:}\;
    $\bmv   =  \widetilde{\bmK}^\transpose \bmu/ \|\widetilde{\bmK}^\transpose \bmu\|_2$\;
    $\bmu   =  \widetilde{\bmK} \bmv/ \|\widetilde{\bmK} \bmv\|_2$\;
    \nonl \emph{Spectral normalization:}\;
    $\widetilde{\bmK}  = \widetilde{\bmK}/(\bmu^\transpose (\widetilde{\bmK} \bmv) +\varepsilon)$\;
    \nonl \emph{Stable rank ($\beta < 1$):}\;
    $\bmS_1 = \bmu\bmv^\transpose$\;
    $\bmS_2 = \widetilde{\bmK} - \bmS_1$\;
    $\gamma = \sqrt{\beta m - 1}/ \|\bmS_2\|_F$\;
    \If{$\gamma < 1$}{
    \nonl $\widetilde{\bmK} = \bmS_1 + \gamma\bmS_2$
    }
    \mbox{$\widetilde{\bmK} = \text{Reshape}(\widetilde{\bmK}^\transpose,[k,k,m,m])$}\\
    \nonl \emph{Training step:}\;
    $\bmK  = \bmK - \eta \nabla_\bmK L(\alpha \,\widetilde{\bmK})$
   }
}
\caption{\emph{SRN--$\alpha$--$\beta$} (Sanyal et al. (2020)~\cite{sanyal2020stable})} \label{alg-SRN}
\end{algorithm}
\end{minipage}
\hfill
\begin{minipage}[t]{0.49\textwidth}
\removelatexerror
\footnotesize
\begin{algorithm}[H]
\DontPrintSemicolon
\KwIn{Number of iterations $N$, learning rate $\eta$,\;
\hspace{1cm} number of channels $m$, image size $n$,\;
\hspace{1cm} initial $\bmK \in \mathbb{R}^{k\times k\times m \times m}$, initial $\bmu \in \mathbb{R}^{n \times n \times m}$.}
\Parameter{Spectral norm $\alpha$, stable rank $\beta$.}
\nonl \Begin{
    \nonl \For{$i=1,\dots, N$}{
    $\widetilde{\bmK} = \bmK$\;
    \nonl \emph{Power iteration:}\;
    $\bmv   =  \widetilde{\bmK}^\transpose \ast \bmu/ \|\widetilde{\bmK}^\transpose \ast \bmu\|_2$\;
    $\bmu   =  \widetilde{\bmK} \ast \bmv/ \|\widetilde{\bmK} \ast \bmv\|_2$\;
    \nonl \emph{Spectral normalization:}\;
    $\widetilde{\bmK}  = \widetilde{\bmK}/(\bmu^\transpose (\widetilde{\bmK} \ast \bmv) +\varepsilon)$\;
    \nonl \emph{Stable rank ($\beta < 1$):}\;
    $\bmS_1 = \nabla_{\widetilde{\bmK}}( \bmu^\transpose (\widetilde{\bmK} \ast \bmv))$\;
    $\bmS_2 = \widetilde{\bmK} - \bmS_1$\;
    $\gamma = \sqrt{\beta m - 1/n^2}/ \|\bmS_2\|_F$\;
    \If{$\gamma < 1$}{
    \nonl $\widetilde{\bmK} = \bmS_1 + \gamma\bmS_2$
    }
    \phantom{abc}\;
    \nonl \emph{Training step:}\;
    $\bmK  = \bmK - \eta \nabla_\bmK L(\alpha \,\widetilde{\bmK})$ 
   }
}
\caption{\emph{SRN-C--$\alpha$--$\beta$} ({\bf Ours})} \label{alg-SRNL}
\end{algorithm}
\end{minipage}
\end{figure*}

\subsection{Stable rank normalization for convolutional layers}

A few methods have been proposed before to enforce the constraints of Approaches~1 and~2. \emph{Spectral normalization} (SN), introduced by Miyato et al.~\cite{miyato2018spectral} and popularized in GAN training~\cite{zhang2019self,brock2019large}, allows one to fix the spectral norm of convolutional layers to a desired value $\alpha$. \emph{Stable rank normalization} (SRN), introduced by Sanyal et al.~\cite{sanyal2020stable}, builds on top of the previous work and allows one to also control the stable rank with a parameter $\beta \in [0,1]$ (Algorithm~\ref{alg-SRN}). However, as observed before in~\cite{scaman2018lipschitz,gouk2018regularisation}, there is an issue with SN and by extension with SRN: they operate on a 2D reshaping of the kernel tensor $\bmK$ instead of operating on the matrix of the convolutional layer $\bmW$ and are therefore unable to enforce stability through the Hard and Soft Lipschitz Constraints, as we verify experimentally in Section~\ref{sec:Constrained Models}. Unfortunately, operating on $\bmW$ directly is impossible: the matrix is too large to be expressed explicitly\footnote{For a kernel size $k$, a number of input and output channels $m$ and an image size $n$, the dimension of $\bmK$ is $[k, k, m, m]$ (typically around $10^3$ parameters) while the dimension of $\bmW$ is $[n n m, n n m]$ (typically around $10^{13}$ sparse parameters).}. In order to solve this intrinsic limitation, we introduce a version of SRN that operates on $\bmW$ indirectly, using $\bmK$. To distinguish between the two versions, we refer to our algorithm as \emph{Stable Rank Normalization for Convolutional layers} or SRN-C (Algorithm~\ref{alg-SRNL}).

The two algorithms are structurally identical---they consist in a power iteration to compute the spectral norm (steps 2,3), a normalization (step 4) and a re-weighting of a rank one matrix $\bmS_1$ and a residual matrix $\bmS_2$ (steps 5, 6, 7, 8)---but they present a number of key differences. In SRN-C, the random vector $\bmu$ has two more dimensions and is the size of a full input feature map ($[1, n, n, m]$). The kernel is not flattened (steps~1, 9). The power iteration is performed using a convolution ($\widetilde{\bmK} \ast \;\cdot$) and a transposed convolution ($\widetilde{\bmK}^\transpose \ast \;\cdot$) as suggested in~\cite{gouk2018regularisation}, based on the observations that:
\begin{equation}
\begin{split}
\bmv   =  \widetilde{\bmW}^\transpose \bmu \quad &\Leftrightarrow \quad \bmv   =  \widetilde{\bmK}^\transpose \ast \bmu \quad \text{(step 2)} \quad \text{and}\\
\bmu   =  \widetilde{\bmW} \bmv \quad &\Leftrightarrow \quad \bmu   =  \widetilde{\bmK} \ast \bmv \quad \text{(step 3)}.
\end{split}
\end{equation}
The spectral normalization is also performed using a convolution (step~4). The rank one matrix $\bmS_1 = \bmu\bmv^\transpose$ is expressed as a 4D kernel tensor through the gradient of $\bmu^\transpose (\widetilde{\bmK} \ast \bmv)$ with respect to $\widetilde{\bmK}$ (step~5), based on the observation that:
\begin{equation}
\bmu \bmv^\transpose = \nabla_{\widetilde{\bmW}}(\text{trace}(\widetilde{\bmW} \bmv \bmu^\transpose)) = \nabla_{\widetilde{\bmW}}(\bmu^\transpose \widetilde{\bmW} \bmv).
\end{equation}
Finally, writing $\|\widetilde{\bmW}\|_F$ explicitly yields $\|\widetilde{\bmW}\|_F = n \|\widetilde{\bmK}\|_F$ and therefore (step 7):
\begin{equation}
\gamma = \frac{\sqrt{\beta n n m - 1}}{n \|\bmS_2\|_F} = \frac{\sqrt{\beta m - 1/n^2}}{\|\bmS_2\|_F}.
\end{equation}

When $\beta = 1$, SRN and SRN-C are equivalent to performing spectral normalization on $\bmK$ and $\bmW$ respectively. 
When $\beta < 1$, they also have an effect on the stable rank of their respective matrices. We found experimentally that SRN multiplies the training time by a factor of $\approx 1.8$ and SRN-C multiplies the training time by a factor of $\approx 2.2$. At inference time, the weights are fixed and normalized convolutions have the same complexity as standard convolutions.

\section{Experiments}
\label{sec:Experiments}

We now illustrate our diagnostic tool on a number of video processing models and show that our stable rank normalization algorithm successfully enforces stability via the Hard and Soft Lipschitz constraints.

\subsection{Unconstrained Models}
\label{sec:Unconstrained Models}

To reflect the variety of architectures used for video processing tasks, we consider two backbone networks and three types of recurrence. The two backbone networks consist in a DnCNN-like~\cite{zhang2017beyond} stack of 10 convolutions and ReLU non-linearities (VDnCNN), and a ResNet-like~\cite{he2016deep} stack of 5 residual blocks containing two convolutions each separated by a ReLU (VResNet). The three types of recurrences are the ones considered in Section~\ref{sec:Motivation}, namely, feature-recurrence~\cite{huang2015bidirectional,chen2016deep,godard2018deep}, frame-recurrence~\cite{frvsr,arias2019kalman} and recurrent latent space propagation (RLSP)~\cite{fuoli2019efficient}. More architectural details are provided in Appendix~C. We focus on video denoising first, and show in Section~\ref{sec:super-resolution} that our main results also apply to video super-resolution.

\setlength{\tabcolsep}{3pt}
\begin{table}
\begin{center}
\caption{Size, processing speed and performance of the different video denoising methods considered, measured on the first frame ($\text{PSNR}_1$), last frame ($\text{PSNR}_7$), and averaged over all the frames ($\text{PSNR}_{\text{mean}}$) on the Vimeo-90k septuplet dataset.}
\label{table:PSNRs}
\begin{tabular}{lccccc}
\hline
\noalign{\smallskip}
  & \# param. & fps & $\text{PSNR}_1$ & $\text{PSNR}_7$ & $\text{PSNR}_{\text{mean}}$ \\
\noalign{\smallskip}
\hline
\noalign{\smallskip}
BM3D~\cite{dabov2007image} & n/a & 2 & 33.86 & 33.83 & 33.85\\
VNLB~\cite{arias2015towards} & n/a & $0.02$ & 35.24 & 35.17 & 35.78\\
FastDVDnet~\cite{tassano2020fastdvdnet} & 2.49M & 7 & \textbf{35.25} & 35.19 & 36.05 \\
FRVSR~\cite{frvsr} & 2.49M & 6 & 34.63 & \textbf{36.83} & \textbf{36.24} \\
DBDNet~\cite{godard2018deep} & 965k & 30 & 34.16 & 35.47 & 35.16\\
VDnCNN-frame & \textbf{375k} & \textbf{70} & 33.94 & 34.84 & 34.68\\
VDnCNN-feat & 741k & 40 & 34.05 & 35.02 & 34.79\\
VDnCNN-RLSP & 410k & 60 & 33.95 & 34.98 & 34.77\\
VResNet-frame & \textbf{375k} & \textbf{70} & 34.23 & 35.47 & 35.21\\
VResNet-feat & 557k & 50 & 34.35 & 35.74 & 35.41\\
VResNet-RLSP & 410k & 60 & 34.25 & 35.80 & 35.42\\
\hline
\end{tabular}
\end{center}
\end{table}

\setlength{\tabcolsep}{3pt}
\begin{table*}[ht]
\centering
\caption{Instabilities in 6 models with 2 backbone architectures and 3 types of recurrences. For each model, we show the performance on the 7th frame of the Vimeo-90k validation dataset ($\text{PSNR}_7$), the $1^{st}$ and $9^{th}$ deciles of the instability onsets on a sequence of about 2h20min ($\infty$ means no instabilities observed). We also show the singular value spectrum averaged over the convolutions of the model, computed as in~\cite{sedghi2019singular}, and the temporal receptive field computed using our method.}
\label{table:unconstrained_models}
\begin{tabular}{|c|ccc|}
\hline
 model & \thead{$\text{PSNR}_7$\\ $1^{st}$ dec.\\ $9^{th}$ dec.} & \thead{Average\\ Singular Value\\Spectrum} & Temporal Receptive Field \\
\hline
\thead{VDnCNN\\--frame} & \thead{$34.84$\\$\boldsymbol{\infty}$\\$\boldsymbol{\infty}$} & \thead{\includegraphics[width=3cm,height=1.7cm]{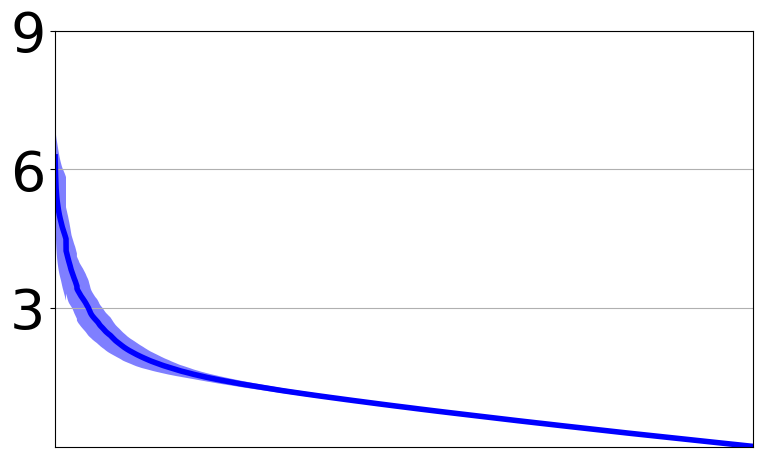}} & \thead{\vspace{-0.55cm}\\ \scriptsize $t$\vspace{0.25cm}\\ \scriptsize $X$\vspace{0.25cm}\\ \scriptsize $Y$} \hspace{-0.3 cm} \thead{\frameNb\\ \includegraphics[width=11cm, height=1.29cm]{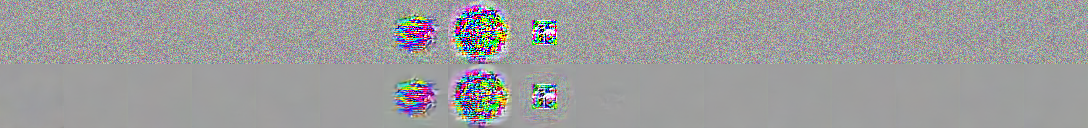}} \vspace{-0.2cm}\\
\thead{VDnCNN\\--feat} & \thead{$35.02$\\$157$\\$5709$} & \thead{\includegraphics[width=3cm,height=1.7cm]{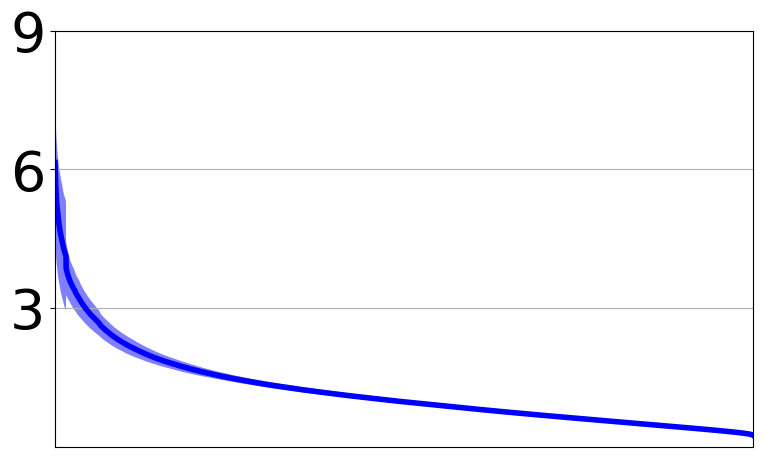}} & \thead{\vspace{-0.55cm}\\ \scriptsize $t$\vspace{0.25cm}\\ \scriptsize $X$\vspace{0.25cm}\\ \scriptsize $Y$} \hspace{-0.3 cm} \thead{\frameNb\\ \includegraphics[width=11cm, height=1.29cm]{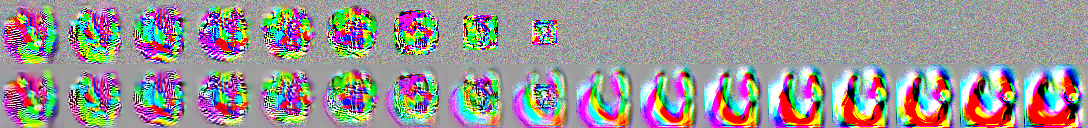}} \vspace{-0.2cm}\\
\thead{VDnCNN\\--RLSP} & \thead{$34.98$\\$74$\\$271$} & \thead{\includegraphics[width=3cm,height=1.7cm]{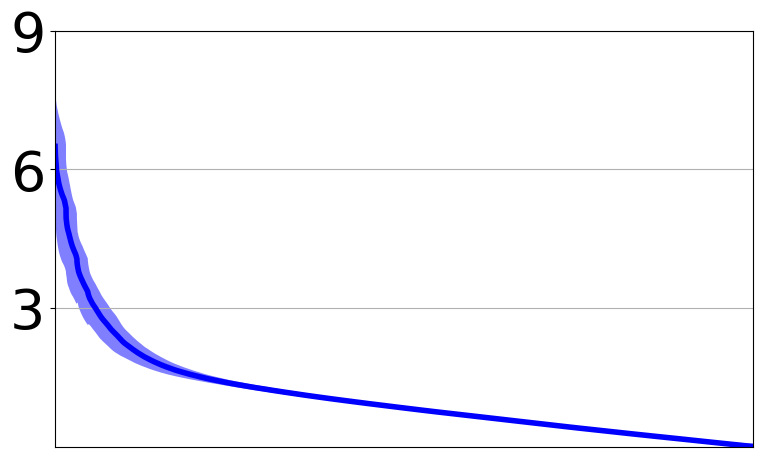}} & \thead{\vspace{-0.55cm}\\ \scriptsize $t$\vspace{0.25cm}\\ \scriptsize $X$\vspace{0.25cm}\\ \scriptsize $Y$} \hspace{-0.3 cm} \thead{\frameNb\\ \includegraphics[width=11cm, height=1.29cm]{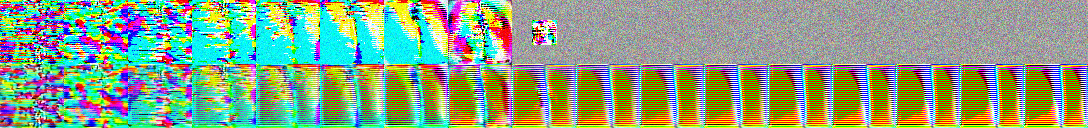}} \vspace{-0.2cm}\\
\thead{VResNet\\--frame} & \thead{$35.47$\\$\boldsymbol{\infty}$\\$\boldsymbol{\infty}$} & \thead{\includegraphics[width=3cm,height=1.7cm]{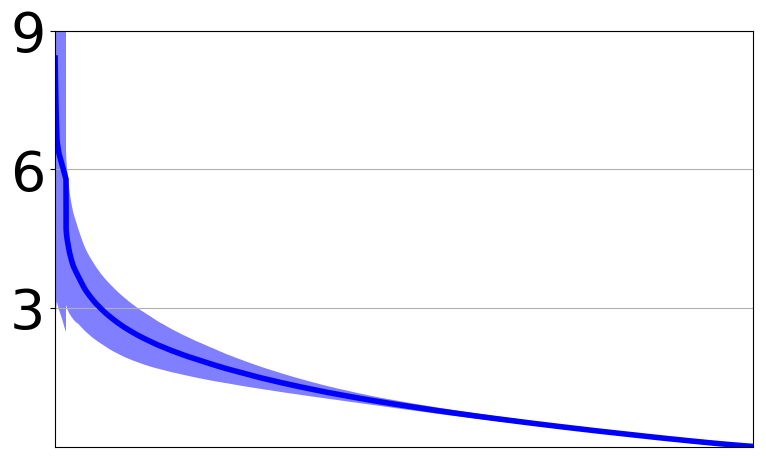}} & \thead{\vspace{-0.55cm}\\ \scriptsize $t$\vspace{0.25cm}\\ \scriptsize $X$\vspace{0.25cm}\\ \scriptsize $Y$} \hspace{-0.3 cm} \thead{\frameNb\\ \includegraphics[width=11cm, height=1.29cm]{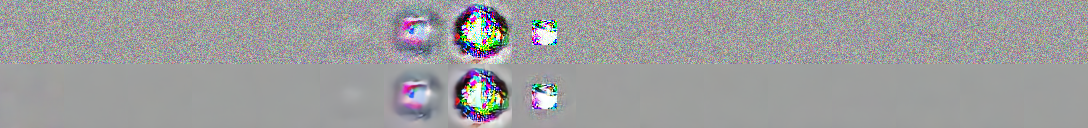}} \vspace{-0.2cm}\\
\thead{VResNet\\--feat} & \thead{$35.74$\\$29$\\$75$} & \thead{\includegraphics[width=3cm,height=1.7cm]{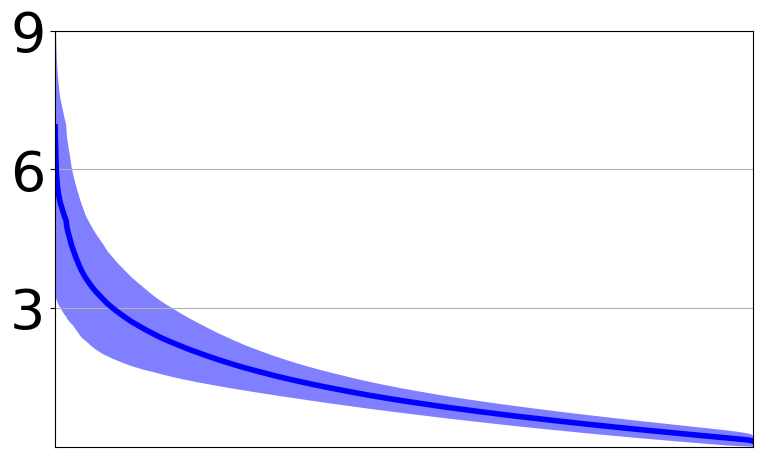}} & \thead{\vspace{-0.55cm}\\ \scriptsize $t$\vspace{0.25cm}\\ \scriptsize $X$\vspace{0.25cm}\\ \scriptsize $Y$} \hspace{-0.3 cm} \thead{\frameNb\\ \includegraphics[width=11cm, height=1.29cm]{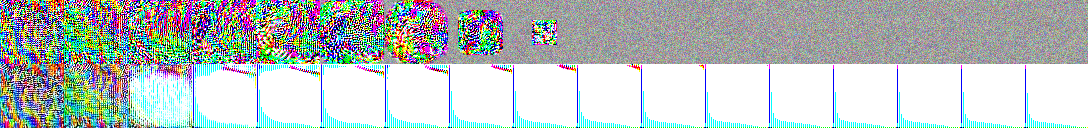}} \vspace{-0.2cm}\\
\thead{VResNet\\--RLSP} & \thead{$35.80$\\$\boldsymbol{\infty}$\\$\boldsymbol{\infty}$} & \thead{\includegraphics[width=3cm,height=1.7cm]{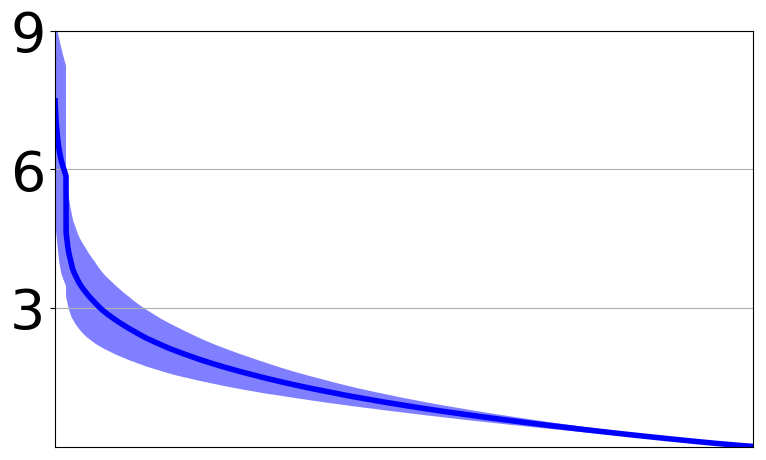}} & \thead{\vspace{-0.55cm}\\ \scriptsize $t$\vspace{0.25cm}\\ \scriptsize $X$\vspace{0.25cm}\\ \scriptsize $Y$} \hspace{-0.3 cm} \thead{\frameNb\\ \includegraphics[width=11cm, height=1.29cm]{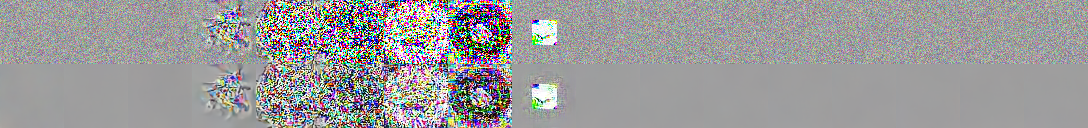}} \\
\hline
\end{tabular}
\end{table*}

We train our models using the Vimeo-90k septuplet dataset \cite{xue2019video}, consisting of about 90k 7-frame RGB sequences with a resolution of $448 \times 256$ downloaded from vimeo.com. We generate clean-noisy training pairs by applying Gaussian noise with standard deviation $\sigma = 30$. The recurrent networks are trained using backpropagation through time on sequences of 7 frames---making use of the full length of the Vimeo-90k sequences---on image crops of $64 \times 64$ pixels. We train using the Adam optimizer with a batch size of 32 for 600k steps. For comparison, we also consider the traditional patch-based methods BM3D~\cite{dabov2007image} and VNLB~\cite{arias2015towards}, the feedforward model FastDVDnet~\cite{tassano2020fastdvdnet} and the recurrent models FRVSR~\cite{frvsr} and DBDNet~\cite{godard2018deep}, which we train in the same conditions as the other recurrent models. In Table~\ref{table:PSNRs}, we show the numbers of parameters and processing speeds\footnote{We use the authors' implementations of BM3D (Matlab) and VNLB (C++). All other networks are implemented in TensorFlow. The processing speeds are indicative of an order of magnitude only.} (fps) of each method, as well as their denoising performances as measured by the PSNR on the first frame ($\text{PSNR}_1$), last frame ($\text{PSNR}_7$) and averaged over all the frames ($\text{PSNR}_{\text{mean}}$) on the first 1024 validation sequences of the Vimeo-90k septuplet dataset. The recurrent architecture FRVSR significantly outperforms other methods on the last frame and in average, while the feedforward architecture FastDVDnet performs best on the first frame. In general, recurrent architectures go through a ``burn-in'' period where performance increases over the first few frames before plateauing to their expected performance on a long sequence. For that reason, we focus on the $\text{PSNR}_7$ metric in the rest of the paper. Our VDnCNN and VResNet models can be considered as simplified versions of FRVSR, with the optical flow alignment network removed and with significantly less parameters. The VResNet backbone systematically outperforms the \mbox{VDnCNN} one, possibly partly because VDnCNN is slower to converge. Frame-recurrent architectures are the lightest and fastest, but feature-recurrence and RLSP yield better performance. Interestingly, VResNet-RLSP performs better than DBDNet~\cite{godard2018deep} ($+0.33$dB on $\text{PSNR}_7$) with about $60\%$ less parameters.

\setlength{\tabcolsep}{3pt}
\begin{table*}[p]
\centering
\caption{SRN and SRN-C with different values of $\alpha$ and $\beta$ on VResNet-feat. The table is organized in the same way as Table~\ref{table:unconstrained_models}.}
\label{table:constrained_models}
\begin{tabular}{|c|ccc|}
\hline
 model & \thead{$\text{PSNR}_7$\\ $1^{st}$ dec.\\ $9^{th}$ dec.} & \thead{Average\\ Singular Value\\Spectrum} & Temporal Receptive Field \\
\hline
\thead{SRN\\$\alpha=1.0$\\$\beta=1.0$} & \thead{$35.64$\\$69$\\$295$} & \thead{\includegraphics[width=3cm,height=1.7cm]{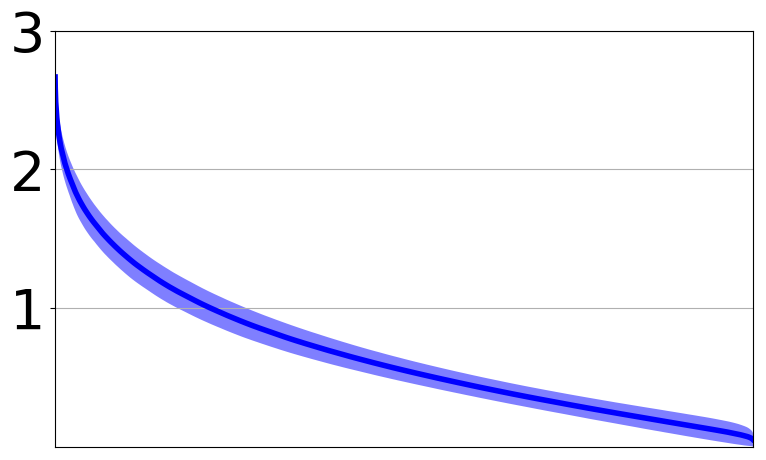}} & \thead{\vspace{-0.55cm}\\ \scriptsize $t$\vspace{0.25cm}\\ \scriptsize $X$\vspace{0.25cm}\\ \scriptsize $Y$} \hspace{-0.3 cm} \thead{\frameNb\\ \includegraphics[width=11cm, height=1.29cm]{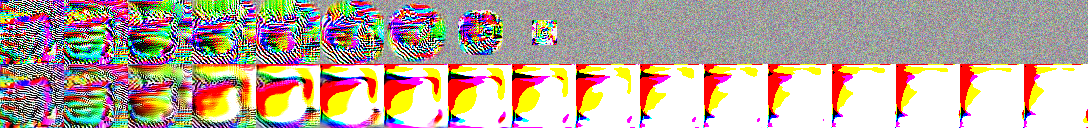}} \vspace{-0.1cm}\\
\hline
\thead{SRN-C\\$\boldsymbol{\alpha=2.0}$\\$\beta=1.0$} & \thead{$35.71$\\$74$\\$264$} & \thead{\includegraphics[width=3cm,height=1.7cm]{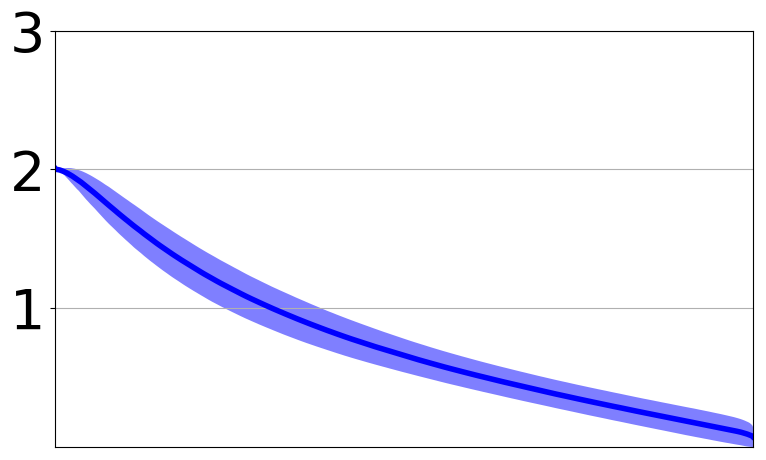}} & \thead{\vspace{-0.55cm}\\ \scriptsize $t$\vspace{0.25cm}\\ \scriptsize $X$\vspace{0.25cm}\\ \scriptsize $Y$} \hspace{-0.3 cm} \thead{\frameNb\\ \includegraphics[width=11cm, height=1.29cm]{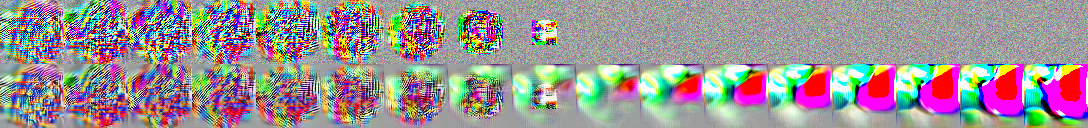}} \vspace{-0.2cm}\\
\thead{SRN-C\\$\boldsymbol{\alpha=1.5}$\\$\beta=1.0$} & \thead{$35.58$\\$84$\\$285$} & \thead{\includegraphics[width=3cm,height=1.7cm]{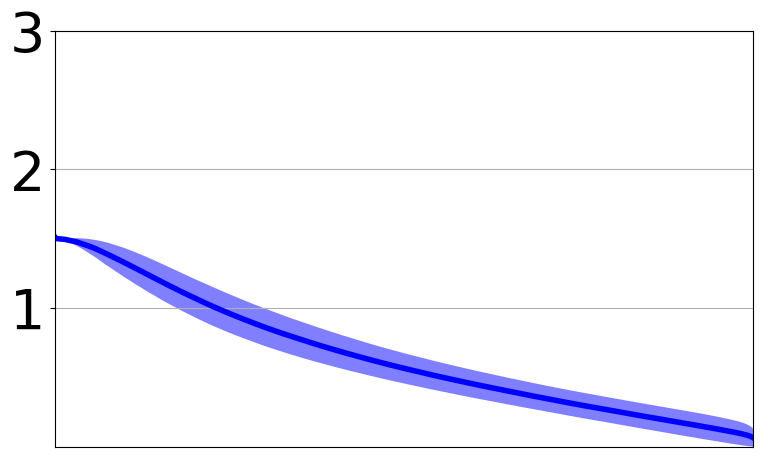}} & \thead{\vspace{-0.55cm}\\ \scriptsize $t$\vspace{0.25cm}\\ \scriptsize $X$\vspace{0.25cm}\\ \scriptsize $Y$} \hspace{-0.3 cm} \thead{\frameNb\\ \includegraphics[width=11cm, height=1.29cm]{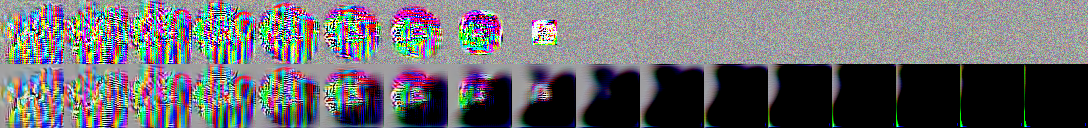}} \vspace{-0.2cm}\\
\thead{SRN-C\\$\boldsymbol{\alpha=1.0}$\\$\beta=1.0$} & \thead{$35.31$\\$\boldsymbol{\infty}$\\$\boldsymbol{\infty}$} & \thead{\includegraphics[width=3cm,height=1.7cm]{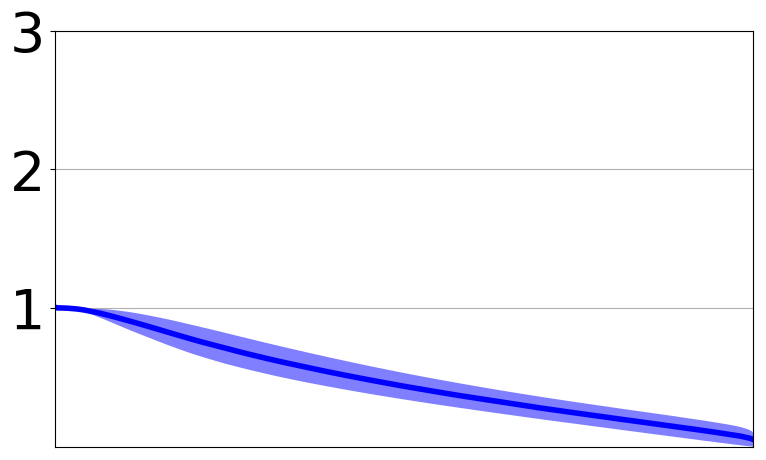}} & \thead{\vspace{-0.55cm}\\ \scriptsize $t$\vspace{0.25cm}\\ \scriptsize $X$\vspace{0.25cm}\\ \scriptsize $Y$} \hspace{-0.3 cm} \thead{\frameNb\\ \includegraphics[width=11cm, height=1.29cm]{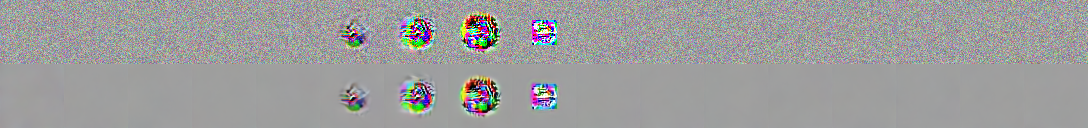}} \vspace{-0.2cm}\\
\thead{SRN-C\\$\boldsymbol{\alpha=0.5}$\\$\beta=1.0$} & \thead{$34.58$\\$\boldsymbol{\infty}$\\$\boldsymbol{\infty}$} & \thead{\includegraphics[width=3cm,height=1.7cm]{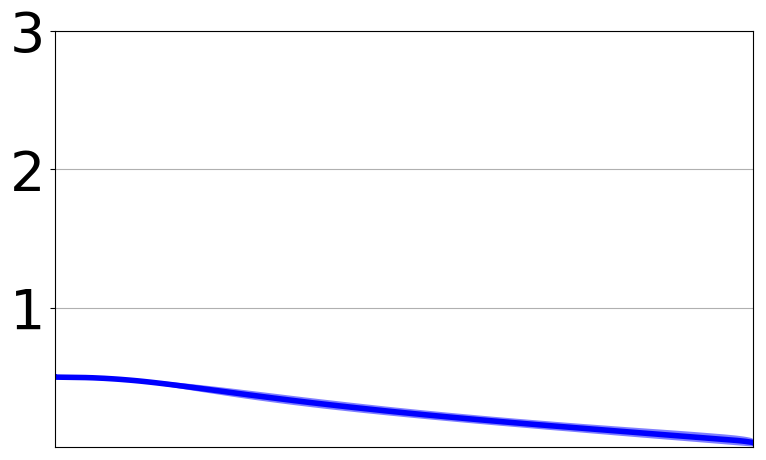}} & \thead{\vspace{-0.55cm}\\ \scriptsize $t$\vspace{0.25cm}\\ \scriptsize $X$\vspace{0.25cm}\\ \scriptsize $Y$} \hspace{-0.3 cm} \thead{\frameNb\\ \includegraphics[width=11cm, height=1.29cm]{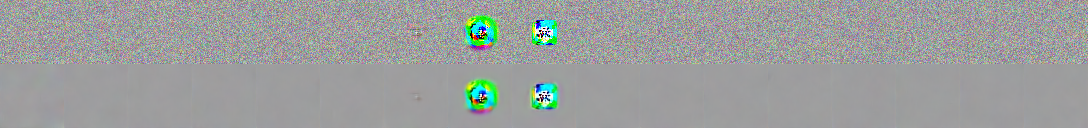}} \vspace{-0.1cm}\\
\hline
\thead{SRN-C\\$\alpha=2.0$\\$\boldsymbol{\beta=0.4}$} & \thead{$35.69$\\$50$\\$258$} & \thead{\includegraphics[width=3cm,height=1.7cm]{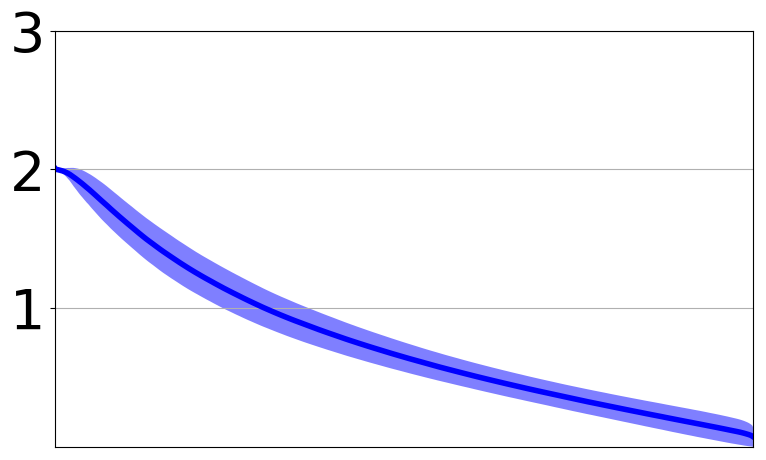}} & \thead{\vspace{-0.55cm}\\ \scriptsize $t$\vspace{0.25cm}\\ \scriptsize $X$\vspace{0.25cm}\\ \scriptsize $Y$} \hspace{-0.3 cm} \thead{\frameNb\\ \includegraphics[width=11cm, height=1.29cm]{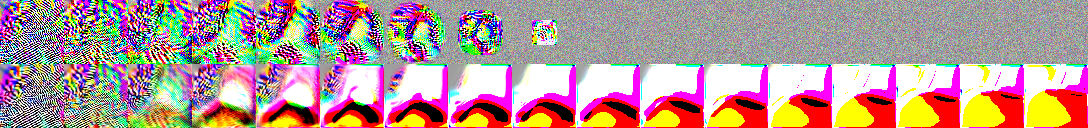}} \vspace{-0.2cm}\\
\thead{SRN-C\\$\alpha=2.0$\\$\boldsymbol{\beta=0.2}$} & \thead{$35.63$\\$26$\\$110$} & \thead{\includegraphics[width=3cm,height=1.7cm]{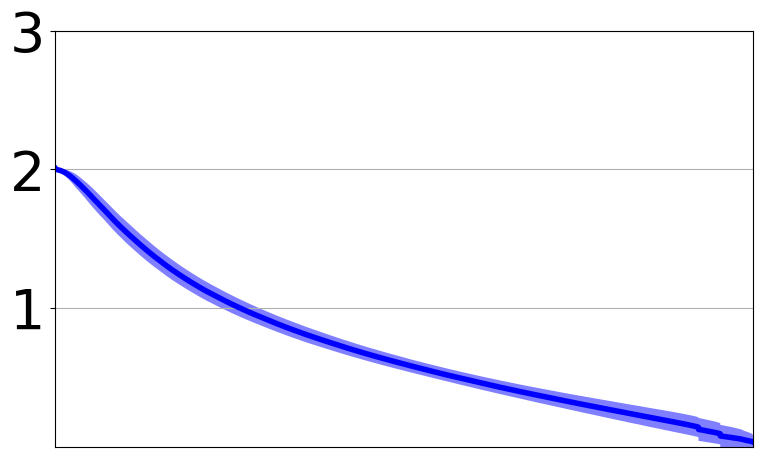}} & \thead{\vspace{-0.55cm}\\ \scriptsize $t$\vspace{0.25cm}\\ \scriptsize $X$\vspace{0.25cm}\\ \scriptsize $Y$} \hspace{-0.3 cm} \thead{\frameNb\\ \includegraphics[width=11cm, height=1.29cm]{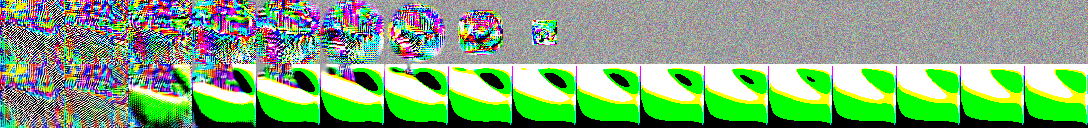}} \vspace{-0.2cm}\\
\thead{SRN-C\\$\alpha=2.0$\\$\boldsymbol{\beta=0.1}$} & \thead{$35.59$\\$\boldsymbol{\infty}$\\$\boldsymbol{\infty}$} & \thead{\includegraphics[width=3cm,height=1.7cm]{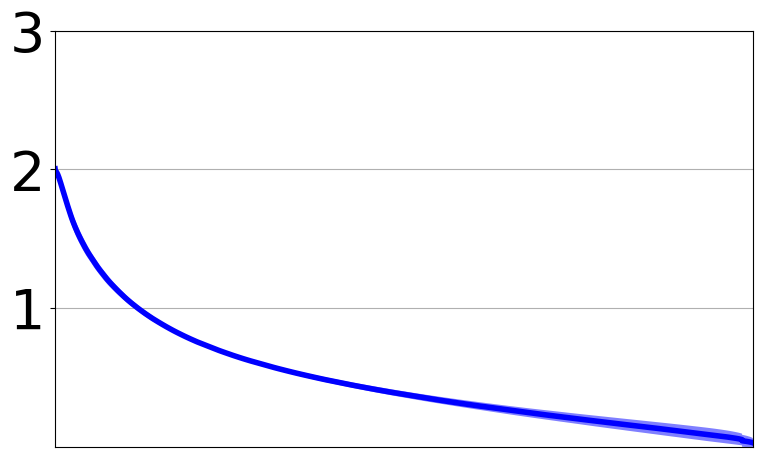}} & \thead{\vspace{-0.55cm}\\ \scriptsize $t$\vspace{0.25cm}\\ \scriptsize $X$\vspace{0.25cm}\\ \scriptsize $Y$} \hspace{-0.3 cm} \thead{\frameNb\\ \includegraphics[width=11cm, height=1.29cm]{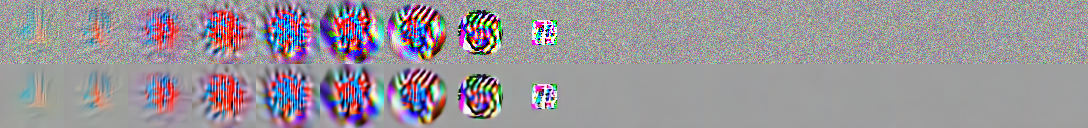}} \vspace{-0.2cm}\\
\thead{SRN-C\\$\alpha=2.0$\\$\boldsymbol{\beta=0.05}$} & \thead{$35.48$\\$\boldsymbol{\infty}$\\$\boldsymbol{\infty}$} & \thead{\includegraphics[width=3cm,height=1.7cm]{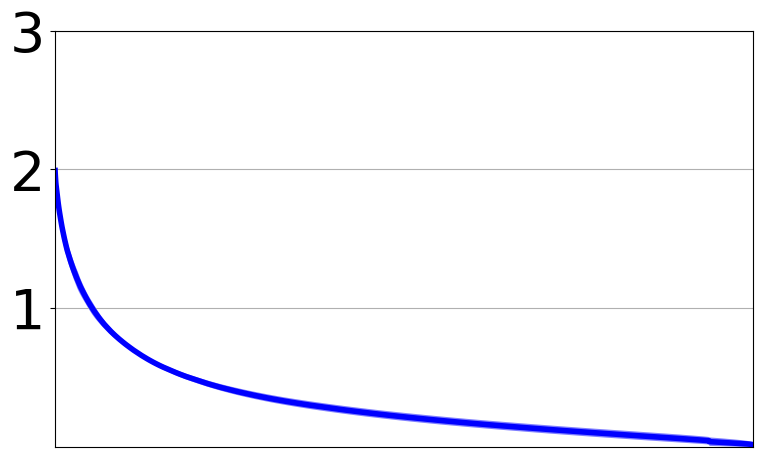}} & \thead{\vspace{-0.55cm}\\ \scriptsize $t$\vspace{0.25cm}\\ \scriptsize $X$\vspace{0.25cm}\\ \scriptsize $Y$} \hspace{-0.3 cm} \thead{\frameNb\\ \includegraphics[width=11cm, height=1.29cm]{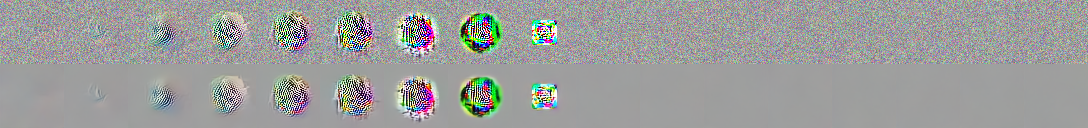}} \\
\hline
\end{tabular}
\end{table*}

\begin{figure*}[p]
\begin{center}
\includegraphics[width=0.97\textwidth]{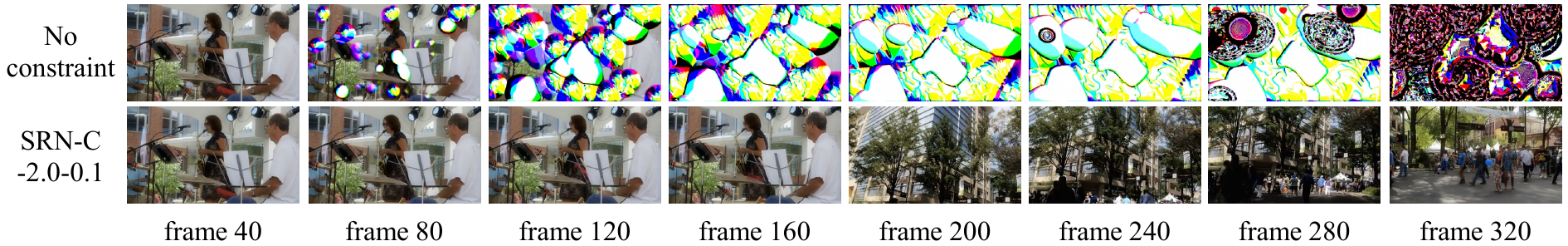}
\end{center}
\vspace{-0.4cm}
\caption{Outputs produced by VResNet-feat in video denoising. Without SRN-C (No constraint), instabilities appear between frame 40 and frame 80 of the chosen video sequence. With SRN-C-2.0-0.1, the outputs are artifact-free on the entire video sequence.}
\label{before-after}
\end{figure*}

To evaluate the stability of each recurrent model, we apply them to one long video sequence lasting approximately 2h20m consisting of several clips downloaded from vimeo.com and concatenated together ($2\times 10^5$ frames). Each time the PSNR drops below 0, we consider that an instability occurs and we \emph{reset} the recurrent features to~0. We call \emph{instability onset} the number of frames leading to an instability. In Table~\ref{table:unconstrained_models}, we report the $1^{st}$ and $9^{th}$ decile of the instability onsets for each model ($\infty$ means no instability observed). According to this test, VDnCNN-frame, VResNet-frame and VResNet-RLSP are stable while VDnCNN-feat, VDnCNN-RLSP and VResNet-feat are unstable. For VDnCNN-feat in particular, the instability onset is above 5709 frames in 10\% of the cases, highlighting the necessity to run this test on very long video sequences. We show examples of output sequences for the 3 unstable models in Appendix~A. In an attempt at explaining why certain models are stable and others are not, we compute the singular value spectra of all their convolutional layers as in~\cite{sedghi2019singular}---this gives us access to their spectral norms or maximum singular value (leftmost value in each spectrum). Unfortunately, the spectra all have comparable profiles and are therefore uninformative, with spectral norms typically around 6: too high to conclude anything about the contractiveness of the recurrent maps using the upper-bound~(\ref{eq:Lipschitz-upper-bound}). Finally, we compute the temporal receptive fields of each model using the approach described in Section~\ref{sec:Diagnosis}. In accordance with the previous test, VDnCNN-feat, VDnCNN-RLSP and VResNet-feat exhibit the characteristic behaviour of unstable models, with long-range temporal dependencies accumulating in the input sequences $X$, resulting in unstable output sequences $Y$ diverging at frame $+40$. The temporal receptive fields of VDnCNN-frame, VResNet-frame and VResNet-RLSP on the other hand, are well-behaved: the information flow is limited to a finite temporal window, input frames in the distant past have no influence on the current frame and future output frames do not diverge. Frame-recurrent models appear to have a short temporal receptive field ($\approx 10$ frames) compared to other models, possibly making this type of recurrence more stable in practice.

\subsection{Constrained models}
\label{sec:Constrained Models}

We saw in the previous section that various models with different backbone architectures and types of recurrence trained in standard conditions are unstable on long video sequences at inference time. We have also discussed in the introduction how the instabilities observed constitute catastrophic failures that are a serious concern for real-world deployment. Now, we show that inference-time stability can be enforced during training, with the help of our \emph{stable rank normalization for convolutional layers} algorithm (SRN-C). We focus on the VResNet-feat architecture, as it appeared to be the most vulnerable to instabilities with 80\% of the onsets happening between 29 and 75 frames only.

First, let us consider the model trained with \mbox{SRN-1.0-1.0} in the first line of Table~\ref{table:constrained_models}. According to the average singular value spectrum, its convolutional layers have spectral norms that are significantly larger than~$1$ at around~$2.5$, and which vary significantly ($\pm 0.2$). This observation confirms that normalizing a 2D reshaping of the convolutional kernel $\bmK$ is a poor approximation of normalizing the convolutional layer $\bmW$: SRN fails to set the spectral norm of $\bmW$ to the desired value $\alpha$, motivating the introduction of \mbox{SRN-C}.

Now, let us consider the models trained with \mbox{SRN-C-$\alpha$-1.0} for $\alpha \in \{2.0, 1.5, 1.0, 0.5\}$ in lines~2 to 5 of Table~\ref{table:constrained_models}. As expected, the spectral norms of all the convolutional layers are now precisely set to their respective values of $\alpha$. Our test on the long video sequence and our temporal receptive field diagnostic then show that the models trained with $\alpha > 1$ are unstable while the models trained with $\alpha < 1$ are stable. This observation confirms that our Hard Lipschitz Constraint is effective at enforcing stability. Interestingly, reducing $\alpha < 1$ shortens the temporal length of the receptive field, a side effect of the recurrence map becoming more contractive. However, reducing $\alpha$ also hurts performance, as measured by the $\text{PSNR}_7$ ($-0.4$dB from $\alpha = 2.0$ to $\alpha = 1.0$), and this motivates the introduction of the Soft Lipschitz Constraint.

Finally, let us consider the models trained with \mbox{SRN-C-2.0-$\beta$} for $\beta \in \{0.4, 0.2, 0.1, 0.05\}$ in lines~6 to 9 of Table~\ref{table:constrained_models}. As expected, varying $\beta$ has no effect on the spectral norm of the convolutional layers, but it has an effect on their stable rank, or the area-under-the-curve of their singular value spectra. Again, our test on the long video sequence and our temporal receptive field diagnostic show that there is a value of $\beta$ for which the stability of the model changes: models trained with $\beta > 0.1$ are unstable while the models trained with $\beta < 0.1$ are stable.  This observation confirms that our Soft Lipschitz Constraint is also effective at promoting stability. Interestingly, reducing $\beta$ also shortens the temporal length of the receptive field but the effect is softer than with $\alpha$, suggesting that controlling the stable rank of the linear layers of a model has a softer effect on its Lipschitz constant than controlling their spectral norms. Importantly, the cost of stability in terms of performance is now lower, and we obtain a stable model that performs better than with the Hard Lipschitz Constraint approach ($+0.18$dB between $\alpha = 1.0, \beta = 1.0$ and $\alpha = 2.0, \beta = 0.1$). We show another illustration of the Soft Lipschitz Constraint with $\alpha = 3.0$ in Appendix~D. Results are summarized in Table~\ref{table:summary}, were we also evaluate each model in terms of the LPIPS metric~\cite{zhang2018unreasonable}, confirming that SRN-C has a neglible impact on the perceptual quality of the outputs. An example of outputs produced by VResNet-feat trained without and with SRN-C is shown in Figure~\ref{before-after}.

\setlength{\tabcolsep}{5pt}
\begin{table}
\begin{center}
\caption{Summary Table. We compare the performances of VResNet-feat stabilised with different variants of SRN-C.}
\label{table:summary}
\begin{tabular}{lccc}
\hline
\noalign{\smallskip}
VResNet-feat with \ldots & $\text{PSNR}_7$
($\uparrow$) & $\text{LPIPS}_7$ ($\downarrow$) &Stable \\
\noalign{\smallskip}
\hline
\noalign{\smallskip}
No Constraint & $35.74$ & $0.080$ & \xmark\\
SRN-C-1.0-1.0 (Hard Cons.) & $35.31$ & $0.079$ & \cmark\\
SRN-C-2.0-0.1 (Soft Cons.) & $35.59$ & $0.083$ & \cmark\\
SRN-C-3.0-0.025 (Soft Cons.) & $35.54$ & $0.075$ & \cmark\\
\hline
\end{tabular}
\end{center}
\end{table}

\subsection{Super-resolution}
\label{sec:super-resolution}

\begin{figure*}[ht]
\begin{center}
\includegraphics[width=0.97\textwidth]{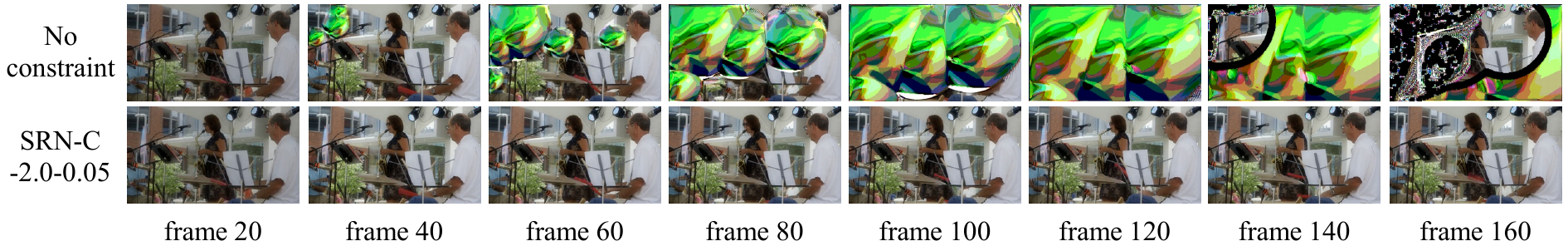}
\end{center}
\vspace{-0.4cm}
\caption{Outputs produced by VResNet-feat in video super-resolution. Without SRN-C (No constraint), instabilities appear between frame 20 and frame 40 of the chosen video sequence. With SRN-C-2.0-0.05, the outputs are artifact-free on the entire video sequence.}
\label{before-after-sr2}
\end{figure*}

To demonstrate that our results generalize to video enhancement tasks other than video denoising, we reproduce here the main results on video super-resolution. We start by training a VResNet-feat model without constraint on the Vimeo-90K dataset for the task of $2\times$ upsampling (a depth-to-space operation is added at the end). We see in Table~\ref{table:video-sr} that the test on the long video sequence and the temporal receptive field diagnostic both confirm that this model is unstable, with 80\% of the instability onsets occuring between 22 and 51 frames only. We then train two more VResNet-feat models, one with a Hard Lipschitz constraint (SRN-C-1.0-1.0), and one with a Soft Lipschitz constraint (SRN-C-2.0-0.05). As expected, both models are now stable according to our two tests. In this case, we do not observe a significant drop of performance for the models trained with constraints, and even observe a slight performance improvement for the model trained with a Soft Lipschitz constraint (+0.06 dB). An example of outputs produced by VResNet-feat trained without and with SRN-C is shown in Figure~\ref{before-after-sr2}.

\setlength{\tabcolsep}{3pt}
\begin{table*}[ht]
\centering
\caption{Video Super-Resolution. The table is organized in the same way as Tables~\ref{table:unconstrained_models} and~\ref{table:constrained_models}.}
\label{table:video-sr}
\begin{tabular}{|c|ccc|}
\hline
 model & \thead{$\text{PSNR}_7$\\ $1^{st}$ dec.\\ $9^{th}$ dec.} & \thead{Average\\ Singular Value\\Spectrum} & Temporal Receptive Field \\
\hline
\thead{No\\ Constraint} & \thead{$32.58$\\$22$\\$51$} & \thead{\includegraphics[width=3cm,height=1.7cm]{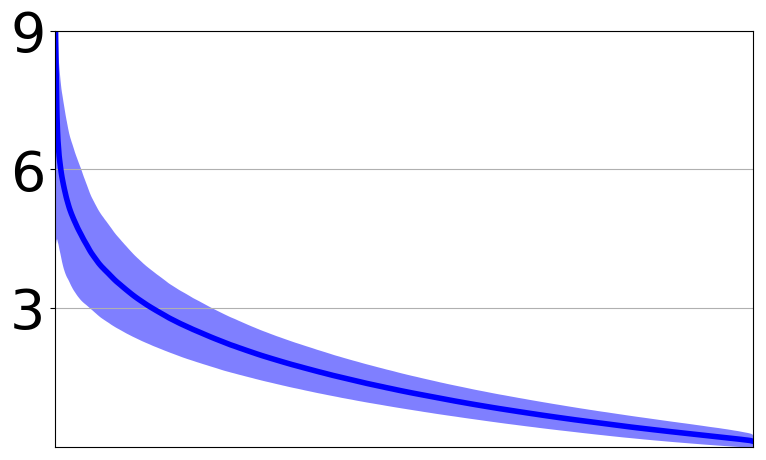}} & \thead{\vspace{-0.55cm}\\ \scriptsize $t$\vspace{0.25cm}\\ \scriptsize $X$\vspace{0.25cm}\\ \scriptsize $Y$} \hspace{-0.3 cm} \thead{\frameNb\\ \includegraphics[width=11cm, height=1.29cm]{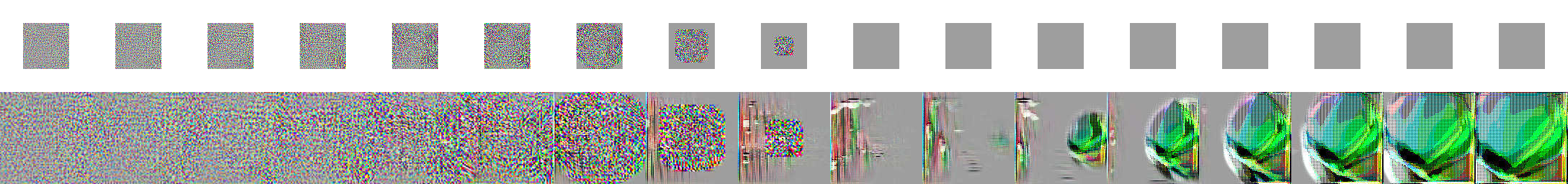}} \vspace{-0.2cm}\\
\thead{SRN-C\\$\boldsymbol{\alpha=1.0}$\\$\beta=1.0$} & \thead{$32.57$\\$\boldsymbol{\infty}$\\$\boldsymbol{\infty}$} & \thead{\includegraphics[width=3cm,height=1.7cm]{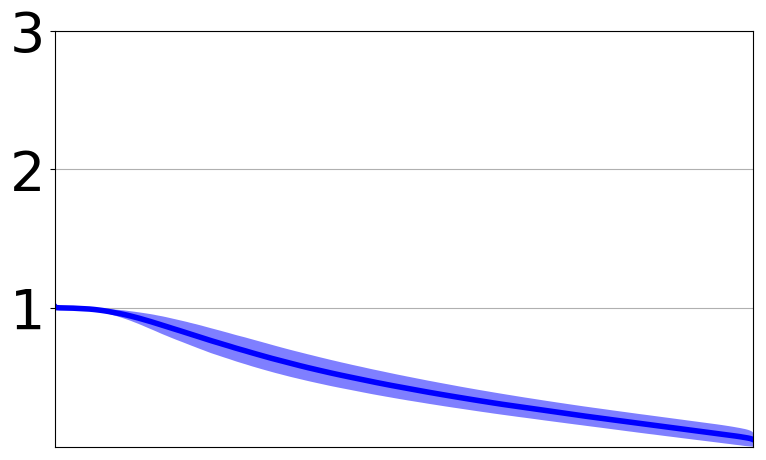}} & \thead{\vspace{-0.55cm}\\ \scriptsize $t$\vspace{0.25cm}\\ \scriptsize $X$\vspace{0.25cm}\\ \scriptsize $Y$} \hspace{-0.3 cm} \thead{\frameNb\\ \includegraphics[width=11cm, height=1.29cm]{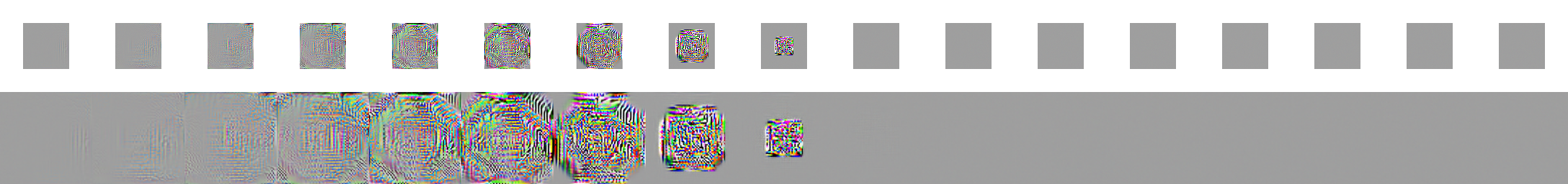}} \vspace{-0.2cm}\\
\thead{SRN-C\\$\alpha=2.0$\\$\boldsymbol{\beta=0.05}$} & \thead{$32.64$\\$\boldsymbol{\infty}$\\$\boldsymbol{\infty}$} & \thead{\includegraphics[width=3cm,height=1.7cm]{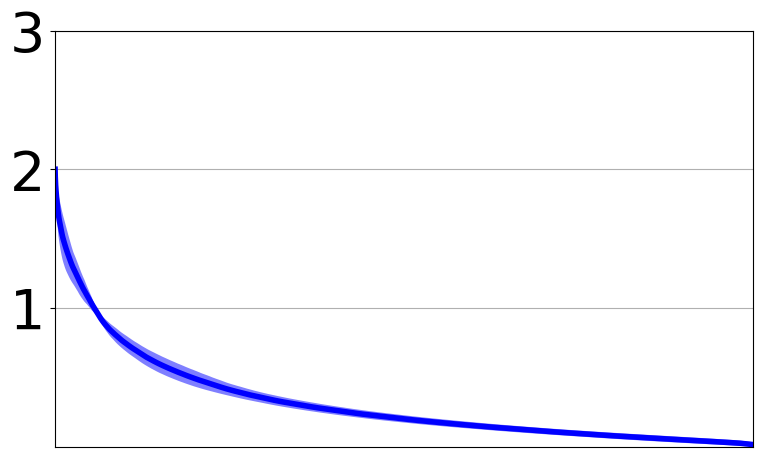}} & \thead{\vspace{-0.55cm}\\ \scriptsize $t$\vspace{0.25cm}\\ \scriptsize $X$\vspace{0.25cm}\\ \scriptsize $Y$} \hspace{-0.3 cm} \thead{\frameNb\\ \includegraphics[width=11cm, height=1.29cm]{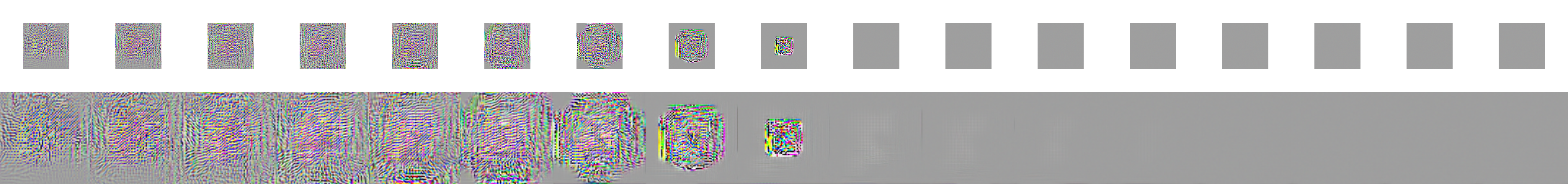}} \\
\hline
\end{tabular}
\end{table*}

\subsection{Discussion}
\label{sec: Discussion}

In the previous sections, we showed that inference-time stability can be enforced during training by constraining the Lipschitz constant of the model to be lower than~1. In this section, we discuss possible alternative strategies.

Given a pre-trained model known to be unstable, one could consider ways to operate it in a stable manner. One approach consists in running the model burst by burst---either without overlap (e.g. running frames 1 to 10, then 11 to 20, then 21 to 30, etc.), or with overlap (e.g. frames 1-10, 6-15, 10-20, etc.)---while resetting the recurrent features to zero between each burst. This strategy prevents instabilities from building up, but it presents a number of issues. Without overlap, the performance fluctuates, the model constantly having to go through a new burn-in period. With overlap, the approach becomes computationally expensive (see Appendix~E). Another approach consists in \emph{dampening the recurrent features} by a factor $\lambda < 1$, allowing a smooth transition between a stable, single frame regime ($\lambda = 0$), and an unstable fully recurrent regime ($\lambda = 1$). This approach is explored in Appendix~F, where we show that the price of stability in terms of performance is much higher than with our Hard and Soft Lipschitz Constraints. 

The instabilities studied in this paper could also be interpreted as a domain adaptation problem. For instance, one hypothesis is that models trained on short sequences fail to generalize to sequences of several hundred frames. However, it is unrealistic to train large recurrent video processing models on sequences of more than 10 to 20 frames---the training process involves backpropagation through time, which has large memory requirements---and even if it was possible, collecting the required data would quickly become impractical. To work around these issues, we perform experiments on a small VDnCNN model where the number of internal convolutions has been reduced to only one, allowing us to unroll the model up to 56 times through time during training, and we generate long sequences with synthetic motion from single frames. We show in Appendix~G that, not only does the model trained on sequences of 56 frames still suffer from instabilities at inference time, but it also suffers from instabilities at training time due to exploding gradients. Another hypothesis is that instabilities are triggered by abrupt scene changes. We show in Appendix~H that in fact, instabilities are more likely to occur on static sequences than on sequences with a lot of scene changes.

\section{Conclusion}

We have identified and characterized a serious vulnerability affecting recurrent networks for video restoration tasks: they can be unstable and fail catastrophically on long video sequences. To avoid problems in practice, we recommend adhering to some guidelines. (1) The stability of the model should always be tested, either by evaluating it on hours of video data, or preferably by actively looking for unstable sequences, using our temporal receptive field diagnostic tool. (2) In safety-critical applications, stability can be guaranteed by applying a Hard Lipschitz Constraint on the spectral norms of the convolutional layers (SRN-C with $\alpha < 1$ and $\beta = 1$). (3) In non safety-critical applications, stability can be obtained with minimal performance loss by applying a Soft Lipschitz Constraint on the stable rank of the convolutional layers (SRN-C with $\alpha > 1$ and $\beta < 1$).

\clearpage
\bibliographystyle{IEEEtran}
\bibliography{biblio}

\begin{IEEEbiography}
    [{\includegraphics[width=1in,height=1.25in,clip,keepaspectratio]{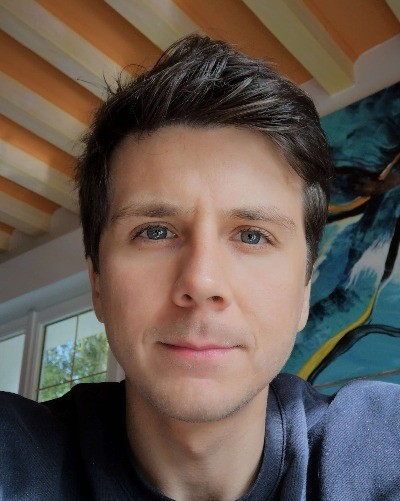}}]{Thomas Tanay}
    is a research scientist at Huawei Technologies Ltd, Noah’s Ark Lab (UK). He received his PhD in Machine Learning after following a MRes in Modelling Biological Complexity at University College London (UK). He previously received a MSc in Cognitive Computing from Goldsmiths, University of London (UK) and a Mechanical Engineering degree from Supm\'eca Paris (France). 
\end{IEEEbiography}

\begin{IEEEbiography}
    [{\includegraphics[width=1in,height=1.25in,clip,keepaspectratio]{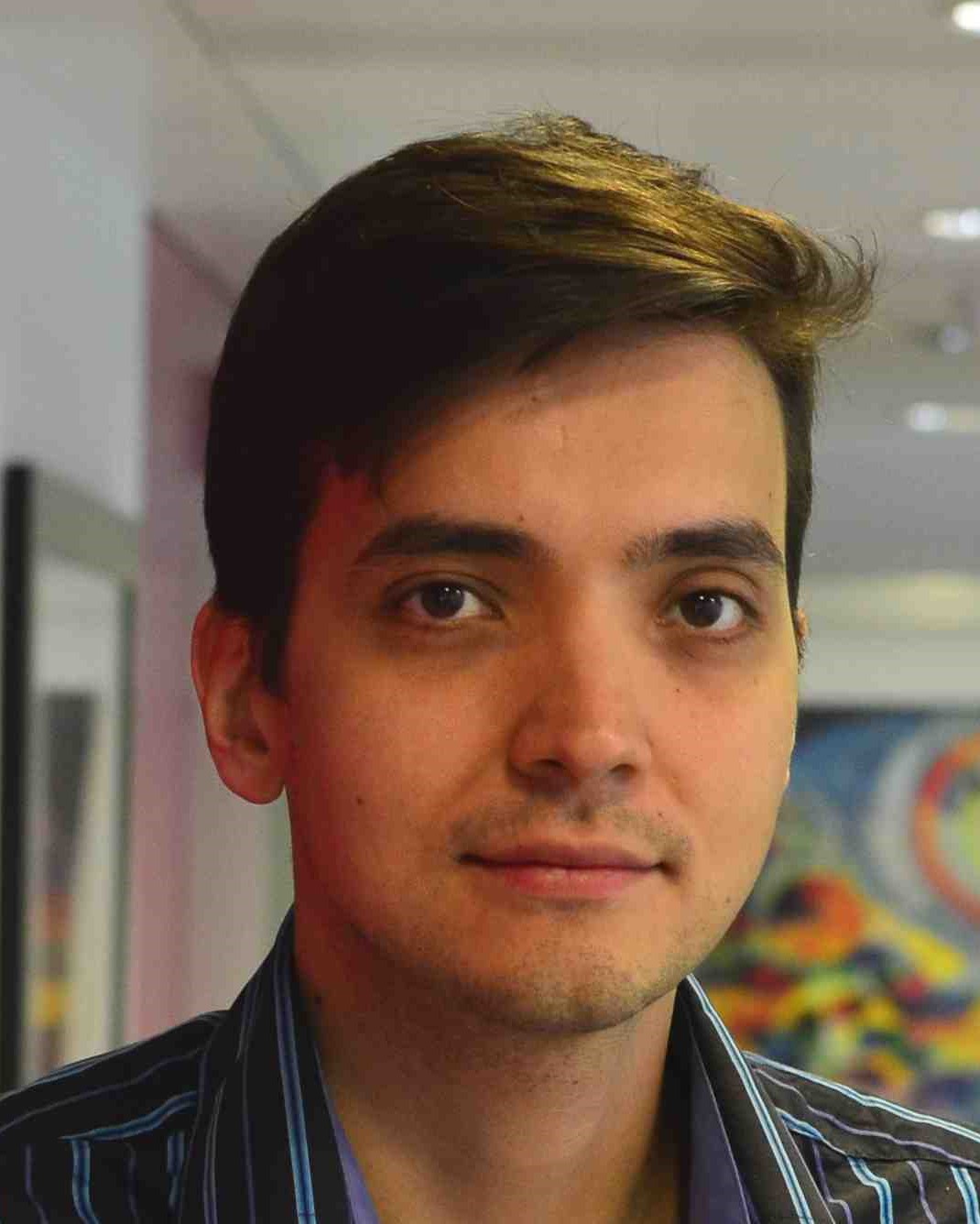}}]{Aivar Sootla}
    is a senior research scientist at Huawei R\&D UK. He has received his MSc in applied mathematics from Lomonosov Moscow State University (Russia) and his PhD in control engineering from Lund University (Sweden). He has held research positions at the Department of Bioengineering,  Imperial College London (UK); at the Montefiore Institute, University of Li\`ege (Belgium); and at the Department of Engineering Science, University of Oxford (UK). 
\end{IEEEbiography}

\begin{IEEEbiography}
    [{\includegraphics[width=1in,height=1.25in,clip,keepaspectratio]{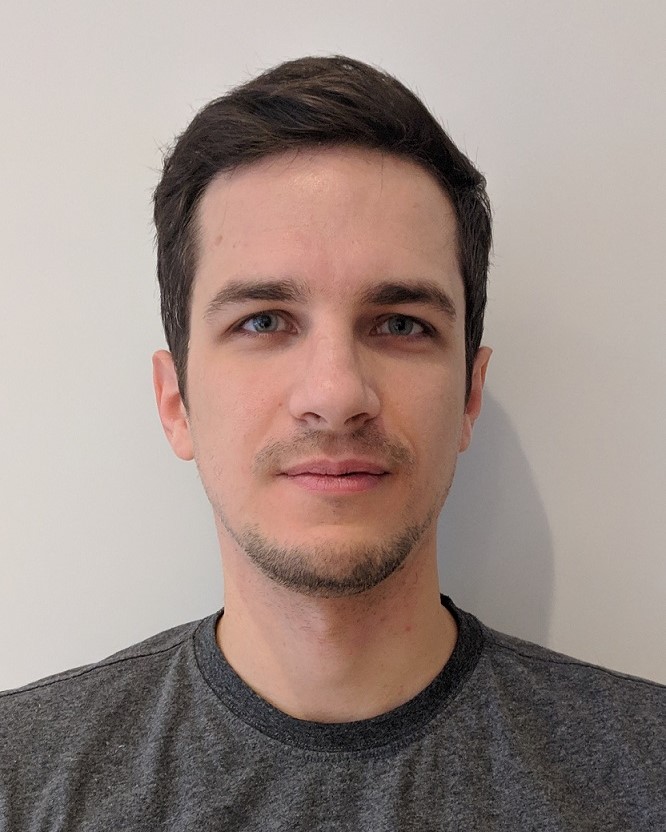}}]{Matteo Maggioni}
received B.Sc. and M.Sc. degrees in computer science from Politecnico di Milano (Italy), a Ph.D. degree in image processing from Tampere University of Technology (Finland), and a Post-Doc with the EEE Department in Imperial College London (UK). He is currently a research scientist with Huawei R\&D (London, UK). His research interests include nonlocal transform-domain filtering, adaptive signal-restoration techniques, and deep learning methods for computer vision and image restoration.
\end{IEEEbiography}

\begin{IEEEbiography}[{\includegraphics[width=1in,height=1.25in,clip,keepaspectratio]{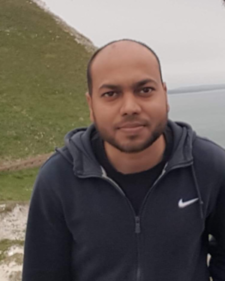}}]{Puneet~K.~Dokania} is a senior researcher at the University of Oxford and a principal research scientist at Five AI (UK). Prior to this, he received his Master of Science in Informatics from Grenoble INP (ENSIMAG), and a Ph.D. in machine learning and applied mathematics from \'Ecole Centrale Paris and INRIA, France. Puneet's current research revolves around developing reliable and efficient algorithms with natural intelligence using deep learning. 
\end{IEEEbiography}

\begin{IEEEbiography}
    [{\includegraphics[width=1in,height=1.25in,clip,keepaspectratio]{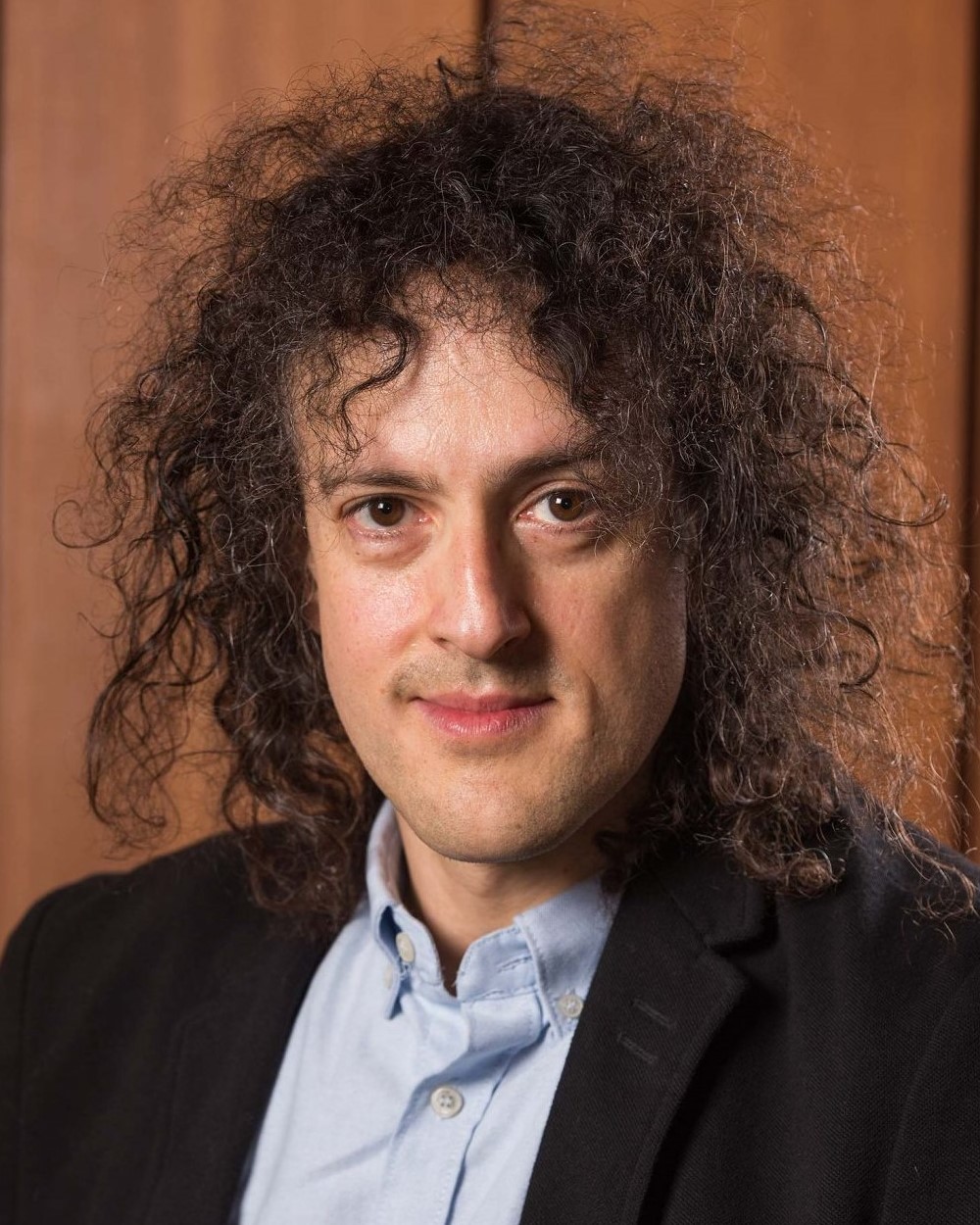}}]{Philip Torr}
did his PhD (DPhil)  at University of Oxford. He left Oxford to work for six years as a research scientist for Microsoft Research, first in Redmond USA in the Vision Technology Group, then in Cambridge UK founding the vision side of the Machine learning and perception group. He then became a Professor in Computer Vision and Machine Learning at Oxford Brookes University. In 2013 he returned to the University of Oxford as full professor. His group has worked mostly in the area of computer vision and machine learning, and has won several awards including the Marr prize, best paper CVPR, best paper ECCV. He is a Fellow of the Royal Academy of Engineering. He has founded companies AIstetic and Oxsight and is Chief Scientific Advisor to FiveAI.
\end{IEEEbiography}

\begin{IEEEbiography}[{\includegraphics[width=1in,height=1.25in,clip,keepaspectratio]{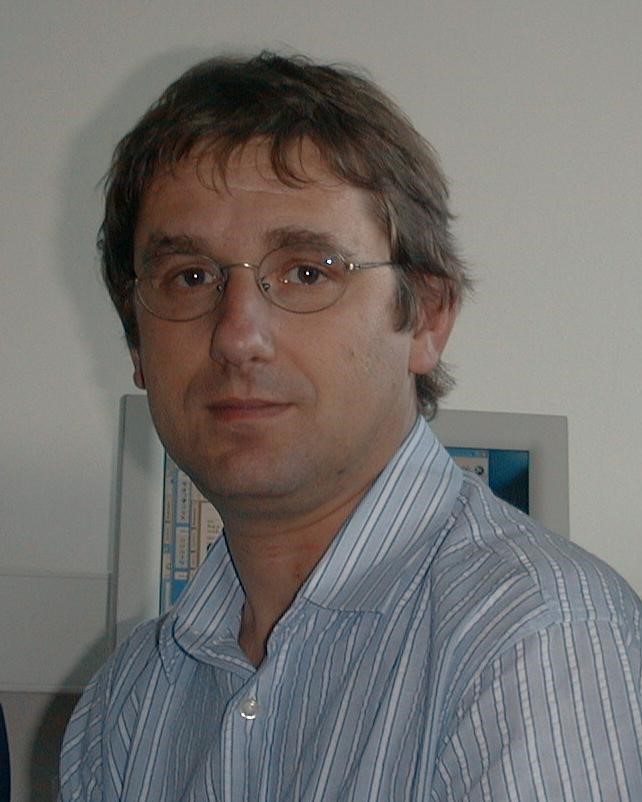}}]{Ale\v{s}~Leonardis} is a Senior Research Scientist and Computer Vision Team Leader at Huawei Technologies Noah’s Ark Lab (London, UK). He is also Chair of Robotics at the School of Computer Science, University of Birmingham and Professor of Computer and Information Science at the University of Ljubljana. Previously, he held research positions at GRASP Laboratory at the University of Pennsylvania and at Vienna University of Technology. He is a Fellow of the IAPR.
\end{IEEEbiography}

\begin{IEEEbiography}
    [{\includegraphics[width=1in,height=1.25in,clip,keepaspectratio]{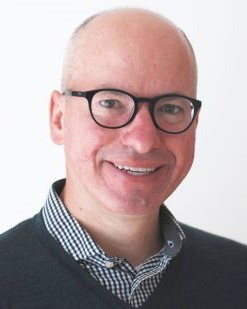}}]{Gregory Slabaugh}
 is Professor of Computer Vision and AI and Director of the Digital Environment Research Institute (DERI) at Queen Mary University of London.  Previous appointments include Huawei, City University of London, Medicsight, and Siemens.  He received the PhD degree in Electrical Engineering from Georgia Institute of Technology.
\end{IEEEbiography}

\newpage
\appendices

\section*{Appendix A: Unstable sequences}

\begin{figure*}[hb]
\begin{center}
\subfloat[VDnCNN-feat \label{instabilities_outputs_VDnCNN-feat}]{\includegraphics[width=0.97\textwidth]{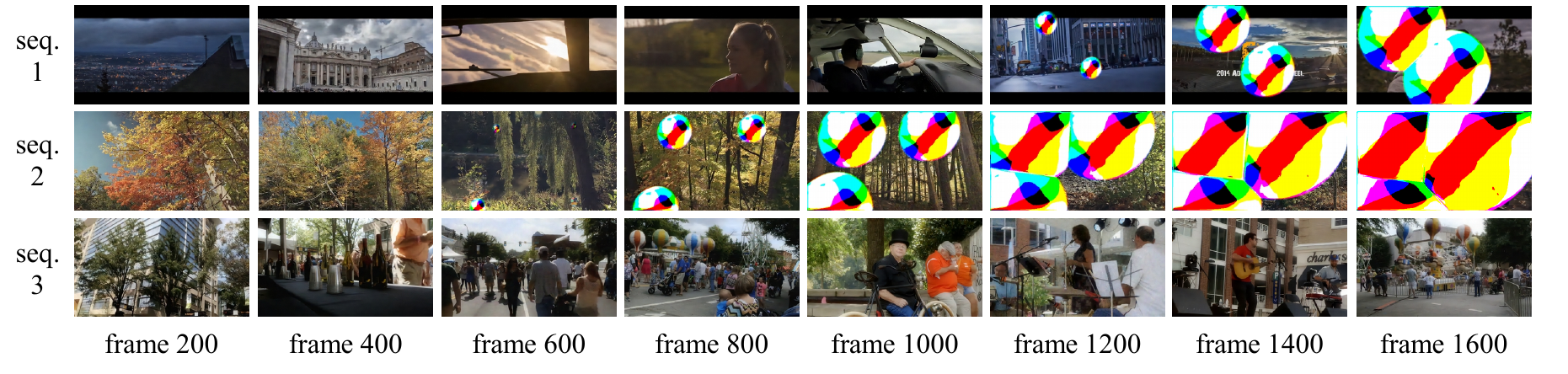}}
\vspace{-0.2cm}

\subfloat[VDnCNN-RLSP \label{instabilities_outputs_VDnCNN-RLSP}]{\includegraphics[width=0.97\textwidth]{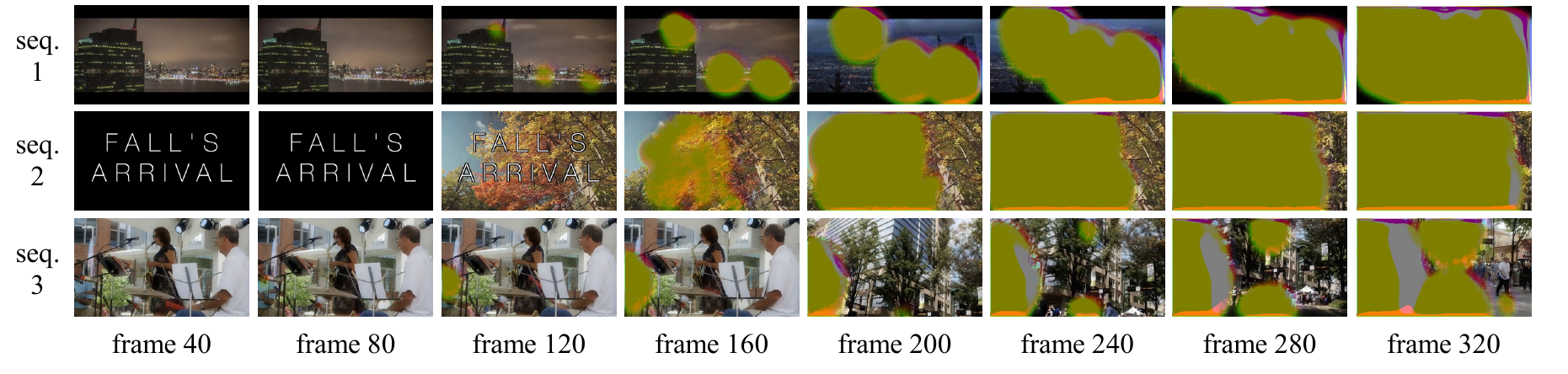}}
\vspace{-0.2cm}

\subfloat[VResNet-feat \label{instabilities_outputs_VResNet-feat}]{\includegraphics[width=0.97\textwidth]{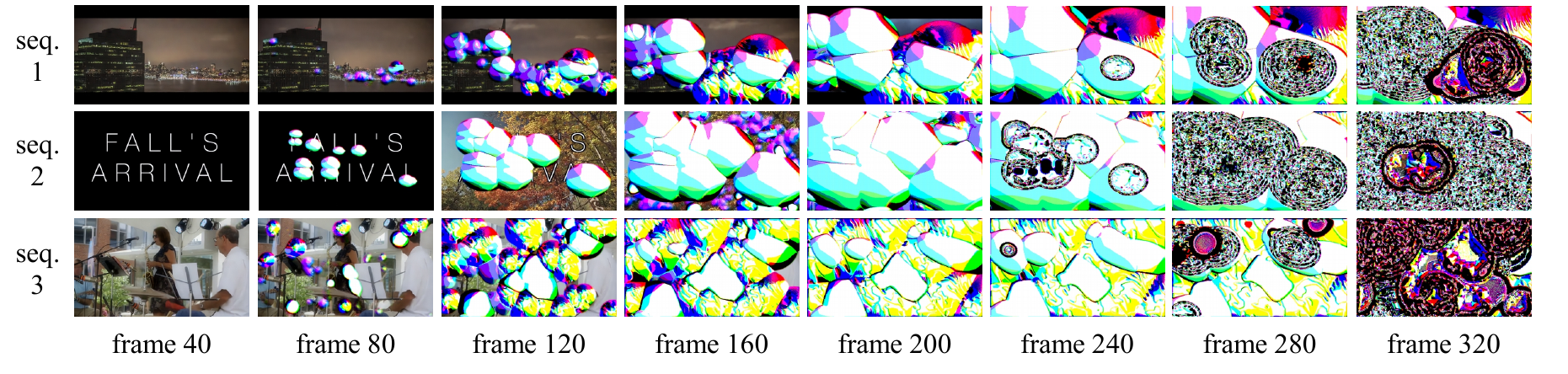}}
\end{center}
\caption{Images generated by the three unstable recurrent video denoisers studied in Section~\ref{sec:Unconstrained Models}, when applied to four sequences of 1600 frames downloaded from vimeo.com. VDnCNN-RLSP and VResNet-feat are unstable on all four sequences after around 120 frames only. VDnCNN-feat is unstable on sequences~1 and~2 after 1200 frames and 800 frames respectively, and it is stable on sequences~3 and~4.}
\label{instabilities_outputs2}
\end{figure*}

We reported in Section~\ref{sec:Unconstrained Models} that VDnCNN-feat, VDnCNN-RLSP and VResNet-feat are unstable on long video sequences at inference time. To illustrate their behaviour in more details, we apply them in Figure~\ref{instabilities_outputs2} to the 3 sequences of 1600 frames already used in Figure~\ref{instabilities}. We can make a number of observations:
\begin{enumerate}
\item Each model is characterized by a distinct ``instability pattern'' that appears at random locations and grows locally until the entire frame is covered.
\item VDnCNN-RLSP and VResNet-feat are unstable on all $3$ sequences after around 120 frames only. VDnCNN-feat is unstable on sequences~1 and~2 after 1200 frames, and it is stable on sequences~3. In this context, it would be easy (but dangerous) to mistake VDnCNN-feat for a stable model.
\item Motion is not necessary to trigger instabilities. The beginning of sequence 2 consists in the static title ``fall's arrival'' and we see that this is enough to trigger an instability on VResNet-feat.
\end{enumerate}

\section*{Appendix B: Feedforward versus recurrent architectures}

Recurrent video processing networks can be unstable at inference time and fail catastrophically on long video sequences. We argue in Section~\ref{sec:Motivation} that this vulnerability is due to the presence of recurrent connections, and that feedforward architectures do not suffer from it. In practice then, what is the motivation for using recurrent architectures over feedforward ones? A first answer is that recurrent architectures perform particularly well in practice, the current state-of-the-art in various video processing applications~\cite{Yang2021NTIRE2C,Son2021NTIRE2C} making heavy use of recurrent connections~\cite{chan2021basicvsr,chan2021basicvsr++}. A more detailed answer is that recurrent processing is particularly adapted to the type of dense information processing over temporally short sequences required for video restoration tasks. To illustrate this, we train VResNet backbones with various temporal connections on Vimeo-90k. \vspace{0.2cm}\\

\begin{figure}[ht]
\begin{center}
\includegraphics[width=0.47\textwidth]{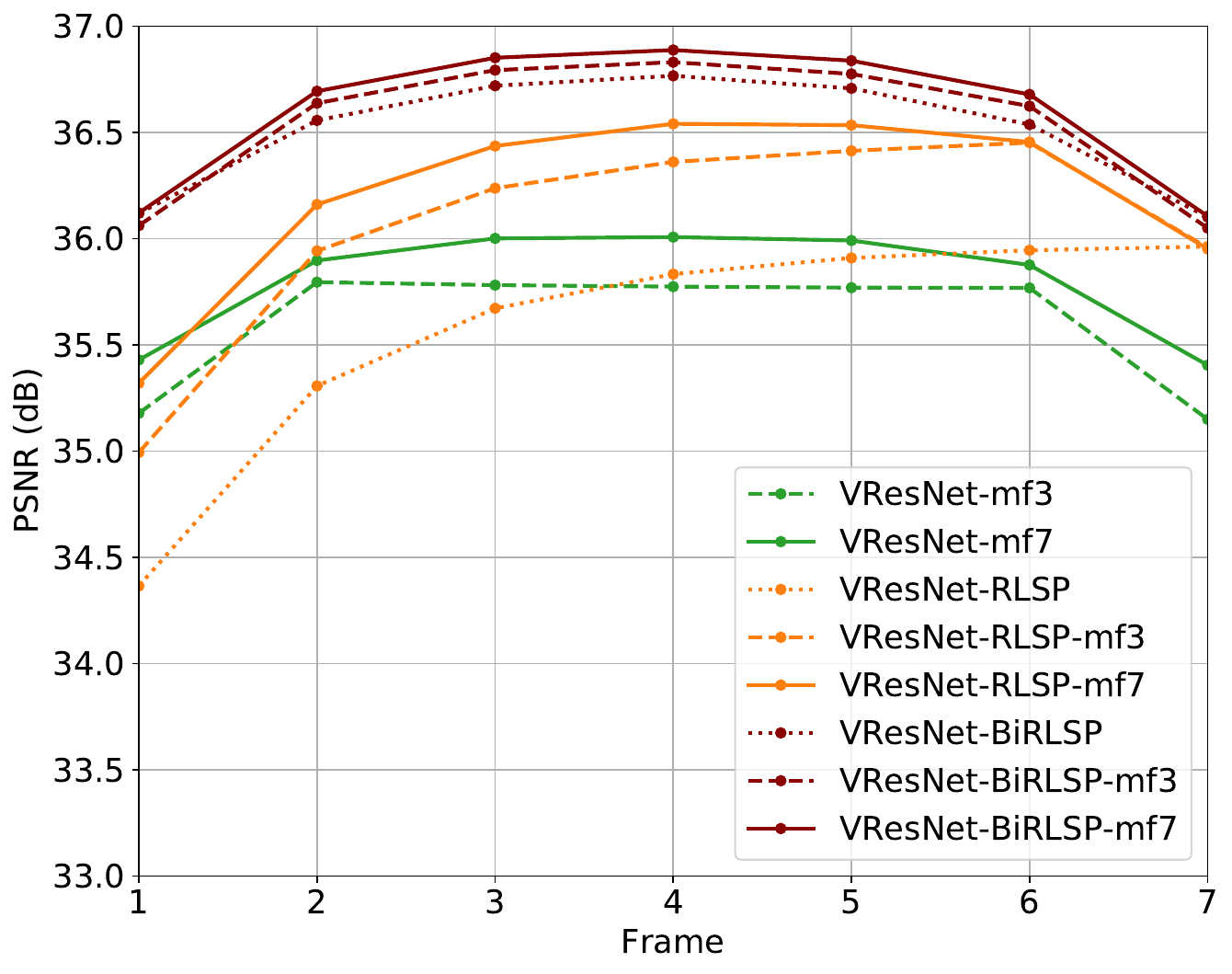}
\end{center}
\caption{PSNR per frame over the Vimeo-90k test set for $\sigma= 30$. Using uni-directional or bi-directional recurrence significantly improves performance over using multi-frame inputs only.}
\label{feedforward_vs_recurrent}
\end{figure}

\noindent We consider two feedforward architectures:
\begin{itemize}[leftmargin=*]
\item \textbf{VResNet-mf3.} Using three consecutive frames as input.
\item \textbf{VResNet-mf7.} Using seven consecutive frames as input.
\end{itemize}
And six recurrent architectures:
\begin{itemize}[leftmargin=*]
\item \textbf{VResNet-RLSP.} Using an RLSP connection.
\item \textbf{VResNet-RLSP-mf3.} Using an RLSP connection and three consecutive frames as input.
\item \textbf{VResNet-RLSP-mf7.} Using an RLSP connection and seven consecutive frames as input.
\item \textbf{VResNet-BiRLSP.} Using an RLSP connection implementing bidirectional recurrence as done in~\cite{huang2015bidirectional,chan2021basicvsr++}: the sequence is processed once in the temporal direction and once in the reverse temporal direction.
\item \textbf{VResNet-BiRLSP-mf3.} The same as above, with three consecutive frames as input.
\item \textbf{VResNet-BiRLSP-mf7.} The same as above, with seven consecutive frames as input.
\end{itemize}
For each model, we then plot the PSNR per frame over the test set in Figure~\ref{feedforward_vs_recurrent}. We see that adding an RLSP connection to feedforward architectures improves performance by about 0.5dB (the computational cost is negligible: the number of input channels to one convolution is simply doubled). Using bidirectional recurrence improves performance by another 0.4dB (although this time the computational cost is doubled). Note that the feedforward architecture VResNet-mf7 and the recurrent architecture VResNet-RLSP-mf7 have access to the same information at all time (the seven input frames), hence the superior performance of VResNet-RLSP-mf7 can only be attributed to recurrent processing.

\section*{Appendix C: Architectural details}

\begin{figure}[ht]
\begin{center}
\subfloat[VDnCNN \label{VDnCNN}]{\includegraphics[width=0.43\textwidth]{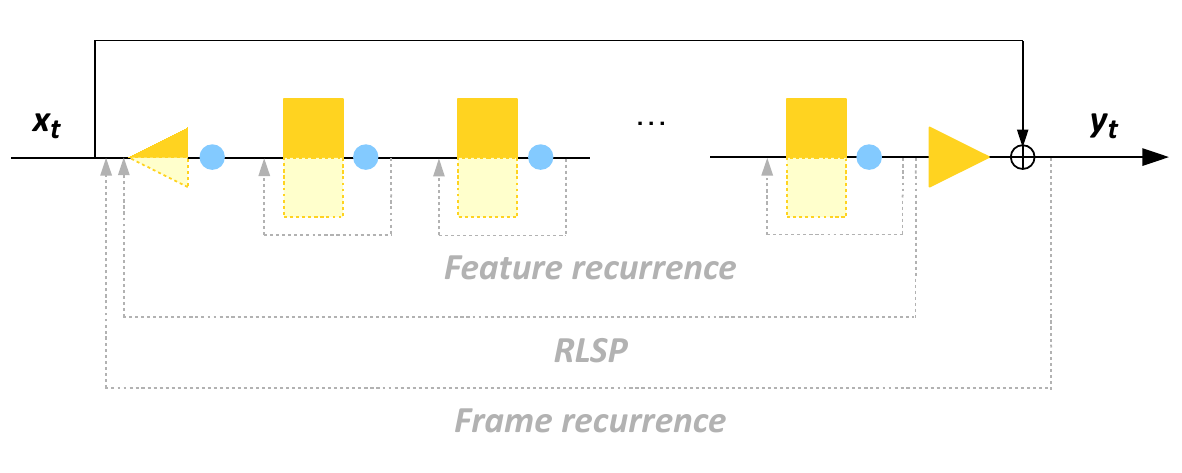}}
\vspace{-0.2cm}

\subfloat[VResNet \label{VResNet}]{\includegraphics[width=0.43\textwidth]{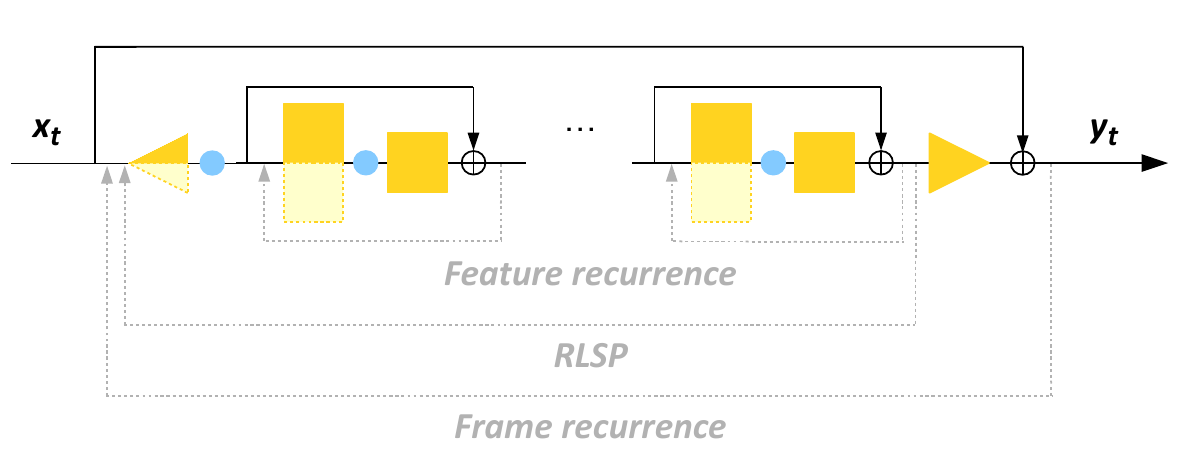}}
\vspace{-0.2cm}

\subfloat{\includegraphics[width=0.45\textwidth]{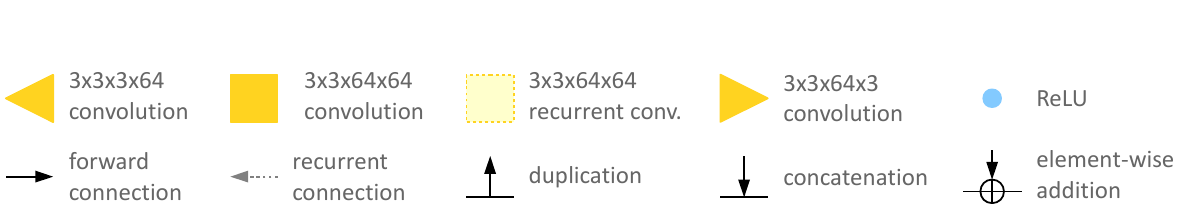}}
\end{center}
\caption{The two architectures and three types of recurrence considered.}
\label{network_architectures}
\end{figure}

The two backbone networks (VDnCNN and VResNet) and the three types of recurrences (feature-recurrence~\cite{huang2015bidirectional,chen2016deep,godard2018deep}, frame-recurrence~\cite{frvsr,arias2019kalman}, RLSP~\cite{fuoli2019efficient}) studied throughout the paper are illustrated in more details in Figure~\ref{network_architectures}.

\newpage
\section*{Appendix D: SRN-C-3.0-$\beta$}

Our SRN-C algorithm allows one to set the spectral norm of a convoluational layer to a desired value $\alpha$ and to control its stable rank with the parameter $\beta \in [0,1]$. For a given model, stability can then be achieved by setting the spectral norms of all the convolutional layers to $1$ (Hard Lipschitz Constraint), or by allowing the spectral norms to be larger than $1$ and by constraining the stable ranks instead (Soft Lipschitz Constraint). We showed in Table~\ref{table:constrained_models} that a stable VResNet-feat model can be obtained by setting $\alpha = 2.0$ and $\beta = 0.1$.  We now show in Table~\ref{table:SRNL-3.0} that a stable VResNet-feat model can also be obtained by setting $\alpha = 3.0$ and $\beta = 0.025$. As expected, increasing $\alpha$ relaxes the stability constraint and needs to be compensated by a smaller value of $\beta$. In terms of $\text{PSNR}_7$, both models perform very similarly: $35.59$ with ($\alpha = 2.0$, $\beta = 0.1$) versus $35.54$ with ($\alpha = 3.0$, $\beta = 0.025$).

\setlength{\tabcolsep}{3pt}
\begin{table*}[ht]
\centering
\caption{SRN-C-3.0-$\beta$ for $\beta \in \{0.4, 0.2, 0.1, 0.05, 0.025\}$. The table is organized in the same way as Tables~\ref{table:unconstrained_models},~\ref{table:constrained_models} and~\ref{table:video-sr}.}
\label{table:SRNL-3.0}
\begin{tabular}{|c|ccc|}
\hline
 model & \thead{$\text{PSNR}_7$\\ $1^{st}$ dec.\\ $9^{th}$ dec.} & \thead{Average\\ Singular Value\\Spectrum} & Temporal Receptive Field \\
\hline
\thead{SRN-C\\$\alpha=3.0$\\$\boldsymbol{\beta=0.4}$} & \thead{$35.75$\\$27$\\$49$} & \thead{\includegraphics[width=3cm,height=1.7cm]{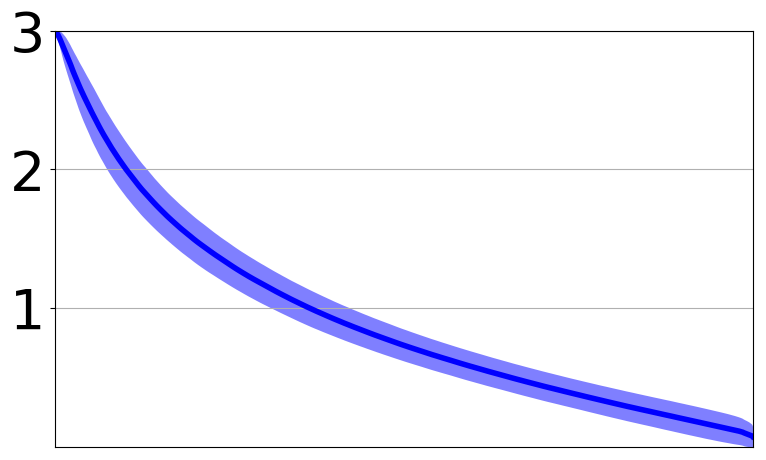}} & \thead{\vspace{-0.55cm}\\ \scriptsize $t$\vspace{0.25cm}\\ \scriptsize $X$\vspace{0.25cm}\\ \scriptsize $Y$} \hspace{-0.3 cm} \thead{\frameNb\\ \includegraphics[width=11cm, height=1.29cm]{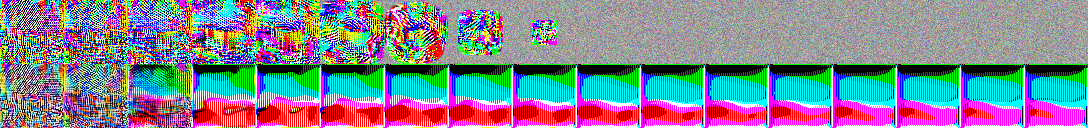}} \vspace{-0.2cm}\\
\thead{SRN-C\\$\alpha=3.0$\\$\boldsymbol{\beta=0.2}$} & \thead{$35.77$\\$51$\\$164$} & \thead{\includegraphics[width=3cm,height=1.7cm]{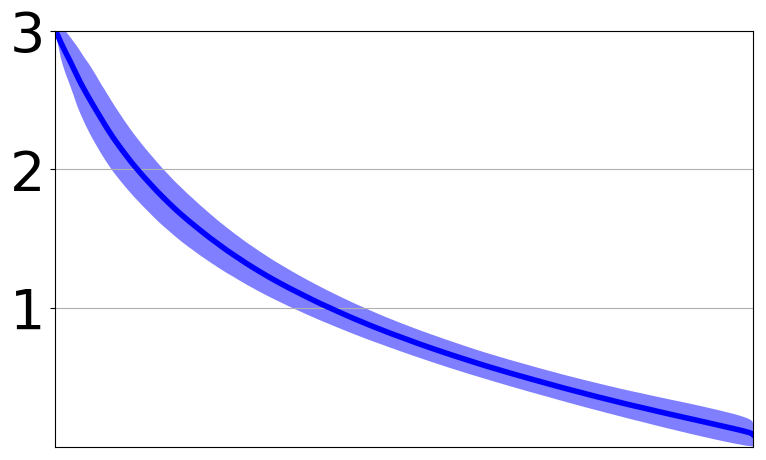}} & \thead{\vspace{-0.55cm}\\ \scriptsize $t$\vspace{0.25cm}\\ \scriptsize $X$\vspace{0.25cm}\\ \scriptsize $Y$} \hspace{-0.3 cm} \thead{\frameNb\\ \includegraphics[width=11cm, height=1.29cm]{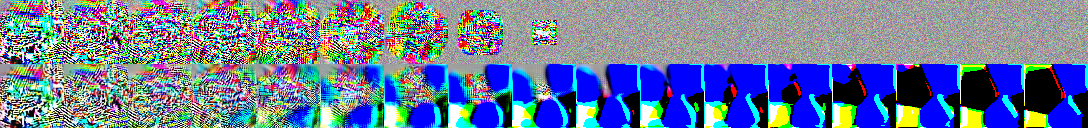}} \vspace{-0.2cm}\\
\thead{SRN-C\\$\alpha=3.0$\\$\boldsymbol{\beta=0.1}$} & \thead{$35.71$\\$48$\\$173$} & \thead{\includegraphics[width=3cm,height=1.7cm]{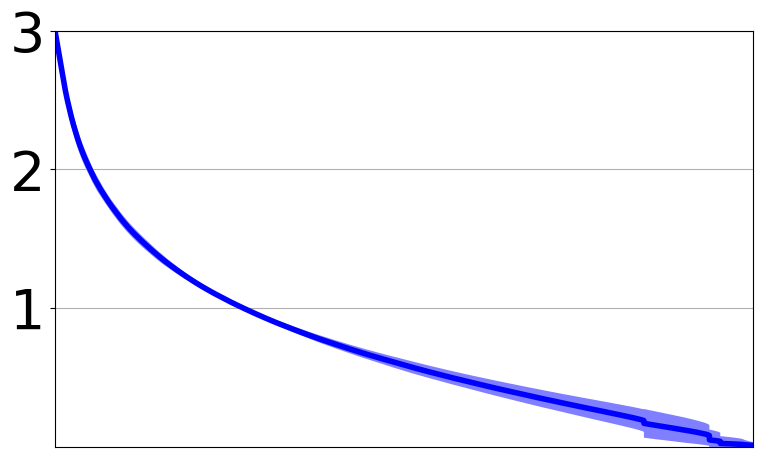}} & \thead{\vspace{-0.55cm}\\ \scriptsize $t$\vspace{0.25cm}\\ \scriptsize $X$\vspace{0.25cm}\\ \scriptsize $Y$} \hspace{-0.3 cm} \thead{\frameNb\\ \includegraphics[width=11cm, height=1.29cm]{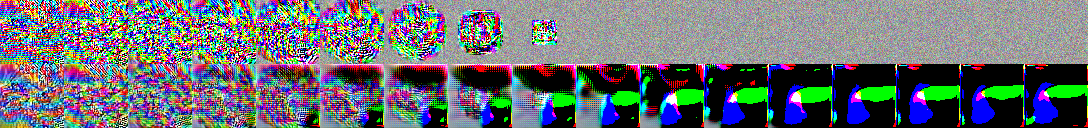}} \vspace{-0.2cm}\\
\thead{SRN-C\\$\alpha=3.0$\\$\boldsymbol{\beta=0.05}$} & \thead{$35.63$\\$62$\\$164$} & \thead{\includegraphics[width=3cm,height=1.7cm]{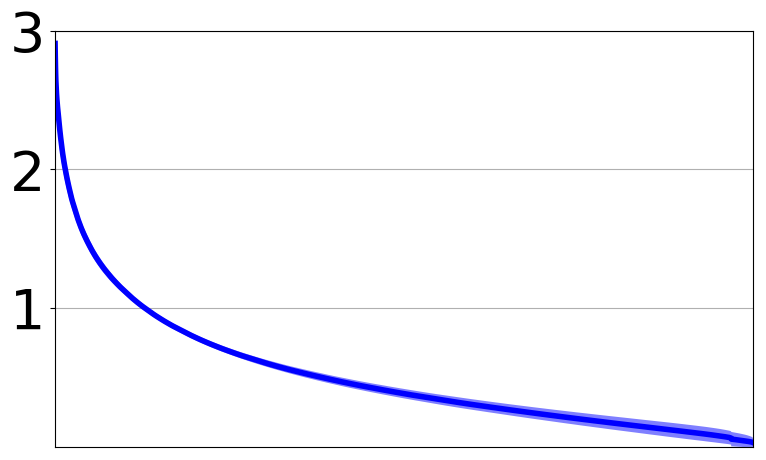}} & \thead{\vspace{-0.55cm}\\ \scriptsize $t$\vspace{0.25cm}\\ \scriptsize $X$\vspace{0.25cm}\\ \scriptsize $Y$} \hspace{-0.3 cm} \thead{\frameNb\\ \includegraphics[width=11cm, height=1.29cm]{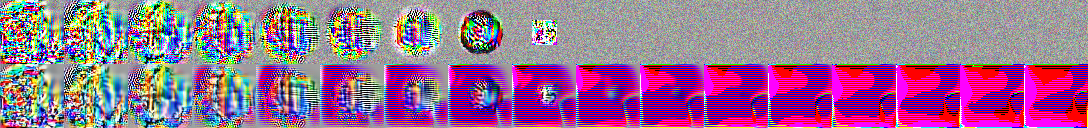}} \vspace{-0.2cm}\\
\thead{SRN-C\\$\alpha=3.0$\\$\boldsymbol{\beta=0.025}$} & \thead{$35.54$\\$\boldsymbol{\infty}$\\$\boldsymbol{\infty}$} & \thead{\includegraphics[width=3cm,height=1.7cm]{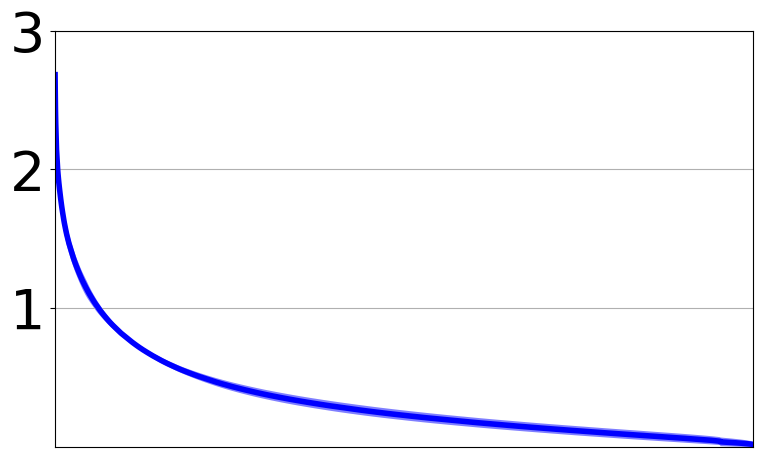}} & \thead{\vspace{-0.55cm}\\ \scriptsize $t$\vspace{0.25cm}\\ \scriptsize $X$\vspace{0.25cm}\\ \scriptsize $Y$} \hspace{-0.3 cm} \thead{\frameNb\\ \includegraphics[width=11cm, height=1.29cm]{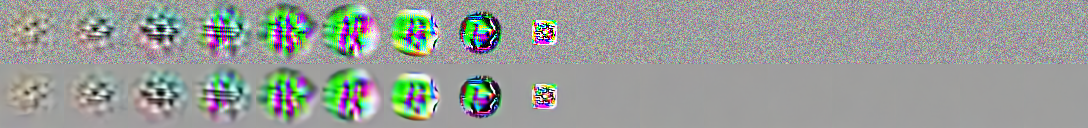}} \\
\hline
\end{tabular}
\end{table*}

\begin{figure*}[ht]
\begin{center}
\includegraphics[width=\textwidth, height=0.33\textwidth]{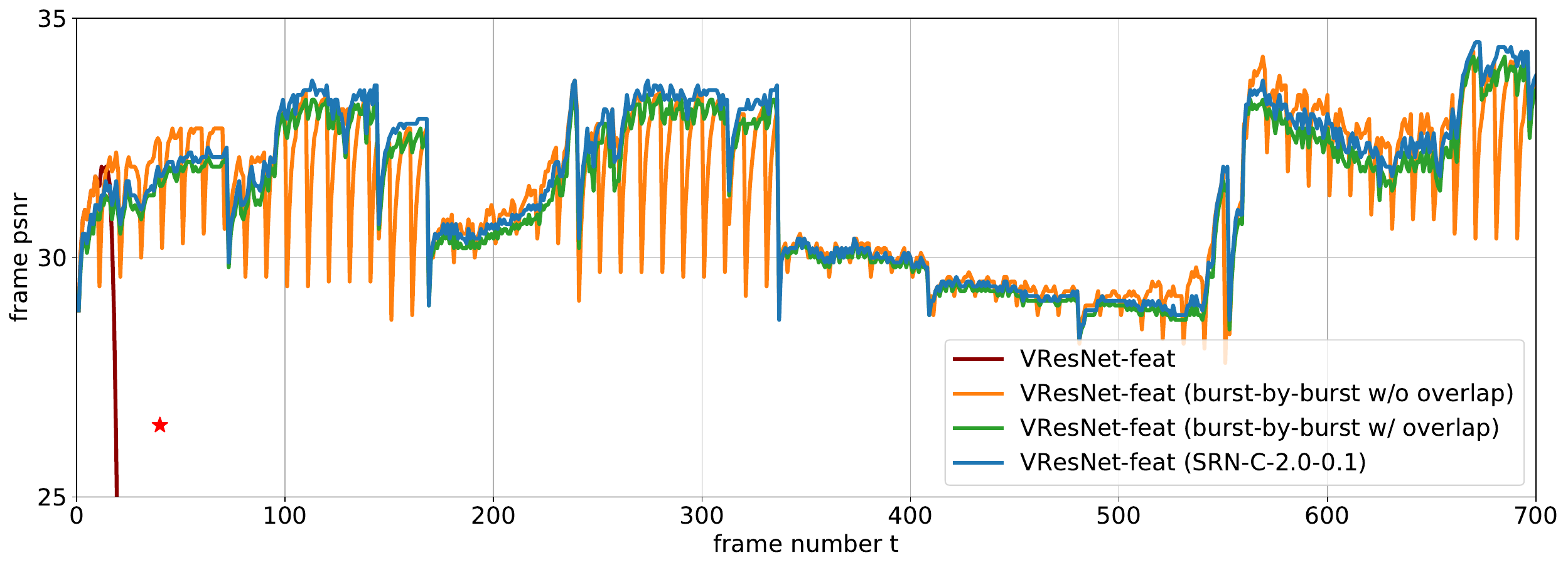}
\end{center}
\caption{Comparison of burst-by-burst processing with a model stabilised with SRN-C on a video sequence of 700 frames. Burst-by-burst processing without overlap results in a fluctuating performance. Burst-by-burst processing with overlap requires to run two models in parallel, each being 50\% smaller to match the computational budget. The model stabilised with SRN-C outperforms the other models on average.}
\label{burst_by_burst}
\end{figure*}

\section*{Appendix E: Burst-by-burst processing}

The instabilities studied in this paper occur at inference time on long video sequences. One simple way to prevent them from occurring is to run recurrent models burst by burst, effectively cutting long video sequences into multiple short ones. Assuming that instabilities never occur on sequences of less than 10 frames for instance, one can run a recurrent model in bursts of 10 frames, resetting the recurrent features to zero at the beginning of each burst. There is a serious drawback with this approach, however: resetting the recurrent features to zero erases all past information fed to the model and the performance can drop by several dBs. A simple solution then consists in running two models burst by burst with an overlap between bursts, only keeping the outputs at the end of each burst. For instance, Model~1 starts at frame 1, Model~2 starts at frame 6, and the output is taken from Model~1 on frames 1-10, 16-20, 26-30, etc., and from Model~2 on frames 11-15, 21-25, 31-35, etc. This approach avoids the performance drops, but it still has drawbacks: matching the computational budget requires to run two models that are 50\% smaller, therefore affecting the overall performance, and the final output alternates between the outputs of two different models, affecting temporal consistency and potentially introducing flickering artifacts. In contrast, enforcing a Soft Lipschitz Constraint into the model during training offers comparable performance without suffering from these drawbacks (see Figure~\ref{burst_by_burst}).

\section*{Appendix F: Feature dampening}

\begin{figure}[ht]
\begin{center}
\includegraphics[width=0.45\textwidth]{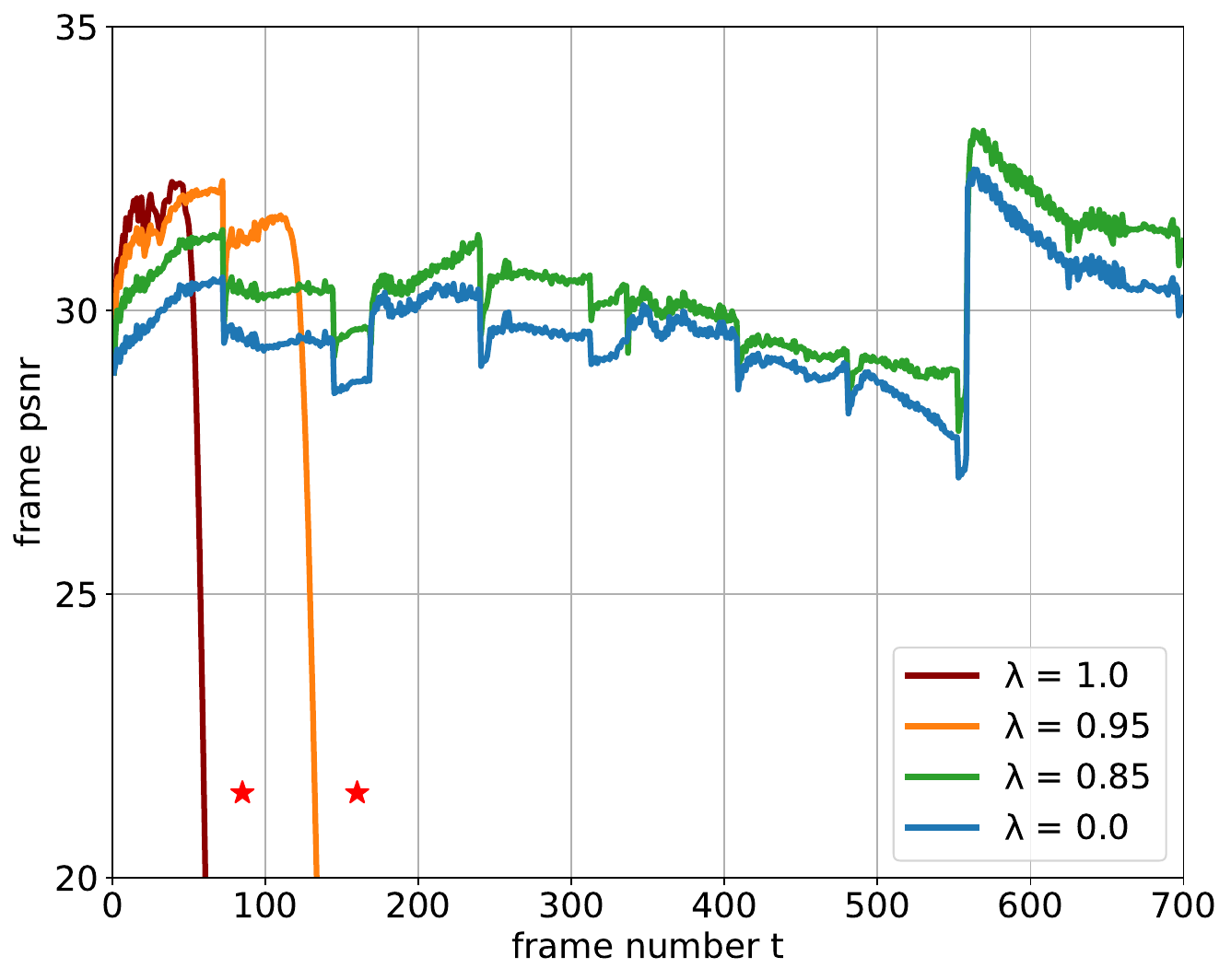}
\end{center}
\vspace{-0.4cm}
\caption{Feature dampening of VResNet-feat for four dampening factors $\lambda$ on a video sequence of 700 frames. Decreasing $\lambda$ improves stability but has a strong negative impact on performance.}
\label{feature_dampening}
\end{figure}

Given a trained model with Lipschitz constant $L$, one brute-force approach to enforce $L < 1$ is to reduce the magnitude of the recurrent weights $\bmK \leftarrow \lambda \, \bmK$ for some $\lambda < 1$. Interestingly, this is equivalent to reducing the magnitude of the recurrent features $h_{t-1} \leftarrow \lambda \, h_{t-1}$ in the convolutions:
\begin{equation*}
    (\lambda \, \bmK) \ast \bmy_{t-1} = \bmK \ast (\lambda \, \bmy_{t-1}).
\end{equation*}
For this reason, we refer to this approach as \emph{feature dampening}. This idea is illustrated on a sequence of 700 frames in Figure~\ref{feature_dampening} for $\lambda \in [1.0, 0.95, 0.85, 0.]$. We see that the model behaves in a stable way for $\lambda = 0.85$. A more detailed study on the number of instabilities measured on a long video sequence and on the temporal receptive fields is provided in Table~5, showing that the model is unstable for $\lambda \in [0.95, 0.85, 0.75, 0.65]$ and only starts to be reliably stable for $\lambda \leq 0.55$ (note that the recurrence is turned off completely for $\lambda = 0.0$). The price to pay in terms of $\text{PSNR}_7$ is high: $34.62$ with $\lambda = 0.55$ versus $35.74$ with $\lambda = 1.0$ ($-1.12$ dB). In comparison, our model trained with SRN-C-2.0-0.1 obtains a $\text{PSNR}_7$ of $35.59$ ($-0.15$ dB).

\setlength{\tabcolsep}{3pt}
\begin{table*}[p]
\centering
\caption{Feature Dampening by a factor $\lambda$ with VResNet-feat. The table is organized in the same way as Tables~\ref{table:unconstrained_models},~\ref{table:constrained_models},~\ref{table:video-sr} and~\ref{table:SRNL-3.0}.}
\label{table:feature_dampening}
\begin{tabular}{|c|ccc|}
\hline
 model & \thead{$\text{PSNR}_7$\\ $1^{st}$ dec.\\ $9^{th}$ dec.} & \thead{Average\\ Singular Value\\Spectrum} & Temporal Receptive Field \\
\hline
\thead{$\lambda = 0.95$} & \thead{$35.31$\\$30$\\$93$} & \thead{\includegraphics[width=3cm,height=1.7cm]{figures/VResNet-feat_sv.png}} & \thead{\vspace{-0.55cm}\\ \scriptsize $t$\vspace{0.25cm}\\ \scriptsize $X$\vspace{0.25cm}\\ \scriptsize $Y$} \hspace{-0.3 cm} \thead{\frameNb\\ \includegraphics[width=11cm, height=1.29cm]{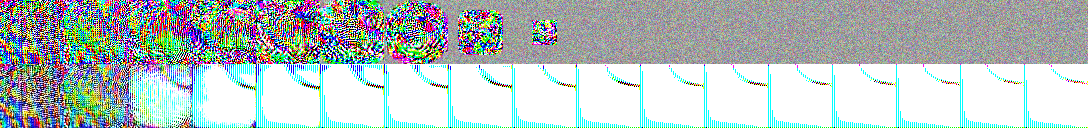}} \vspace{-0.2cm}\\
\thead{$\lambda = 0.85$} & \thead{$34.93$\\$38$\\$257$} & \thead{\includegraphics[width=3cm,height=1.7cm]{figures/VResNet-feat_sv.png}} & \thead{\vspace{-0.55cm}\\ \scriptsize $t$\vspace{0.25cm}\\ \scriptsize $X$\vspace{0.25cm}\\ \scriptsize $Y$} \hspace{-0.3 cm} \thead{\frameNb\\ \includegraphics[width=11cm, height=1.29cm]{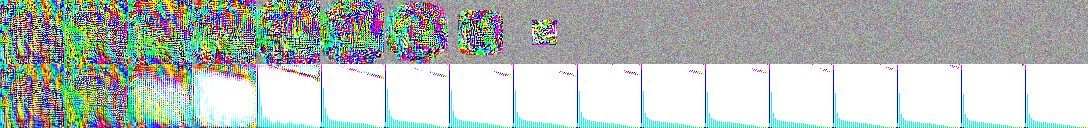}} \vspace{-0.2cm}\\
\thead{$\lambda = 0.75$} & \thead{$34.78$\\$59$\\$1024$} & \thead{\includegraphics[width=3cm,height=1.7cm]{figures/VResNet-feat_sv.png}} & \thead{\vspace{-0.55cm}\\ \scriptsize $t$\vspace{0.25cm}\\ \scriptsize $X$\vspace{0.25cm}\\ \scriptsize $Y$} \hspace{-0.3 cm} \thead{\frameNb\\ \includegraphics[width=11cm, height=1.29cm]{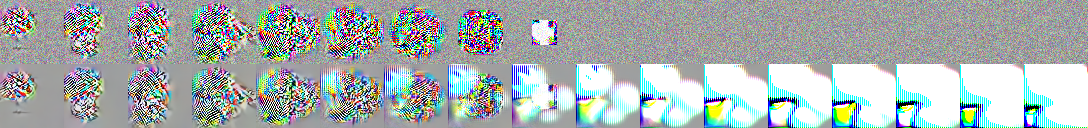}} \vspace{-0.2cm}\\
\thead{$\lambda = 0.65$} & \thead{$34.69$\\$181$\\$38097$} & \thead{\includegraphics[width=3cm,height=1.7cm]{figures/VResNet-feat_sv.png}} & \thead{\vspace{-0.55cm}\\ \scriptsize $t$\vspace{0.25cm}\\ \scriptsize $X$\vspace{0.25cm}\\ \scriptsize $Y$} \hspace{-0.3 cm} \thead{\frameNb\\ \includegraphics[width=11cm, height=1.29cm]{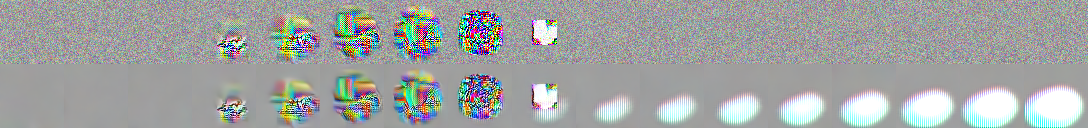}} \vspace{-0.2cm}\\
\thead{$\lambda = 0.55$} & \thead{$34.62$\\$\boldsymbol{\infty}$\\$\boldsymbol{\infty}$} & \thead{\includegraphics[width=3cm,height=1.7cm]{figures/VResNet-feat_sv.png}} & \thead{\vspace{-0.55cm}\\ \scriptsize $t$\vspace{0.25cm}\\ \scriptsize $X$\vspace{0.25cm}\\ \scriptsize $Y$} \hspace{-0.3 cm} \thead{\frameNb\\ \includegraphics[width=11cm, height=1.29cm]{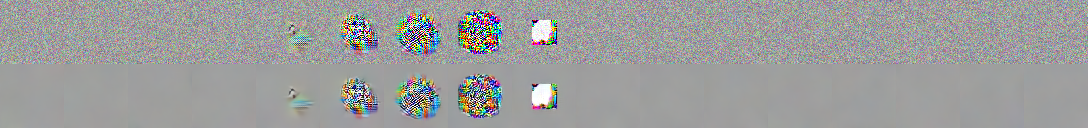}} \vspace{-0.2cm}\\
\thead{$\lambda = 0.45$} & \thead{$34.56$\\$\boldsymbol{\infty}$\\$\boldsymbol{\infty}$} & \thead{\includegraphics[width=3cm,height=1.7cm]{figures/VResNet-feat_sv.png}} & \thead{\vspace{-0.55cm}\\ \scriptsize $t$\vspace{0.25cm}\\ \scriptsize $X$\vspace{0.25cm}\\ \scriptsize $Y$} \hspace{-0.3 cm} \thead{\frameNb\\ \includegraphics[width=11cm, height=1.29cm]{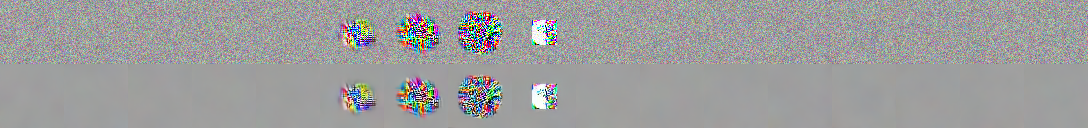}} \vspace{-0.2cm}\\
\thead{$\lambda = 0.0$} & \thead{$34.32$\\$\boldsymbol{\infty}$\\$\boldsymbol{\infty}$} & \thead{\includegraphics[width=3cm,height=1.7cm]{figures/VResNet-feat_sv.png}} & \thead{\vspace{-0.55cm}\\ \scriptsize $t$\vspace{0.25cm}\\ \scriptsize $X$\vspace{0.25cm}\\ \scriptsize $Y$} \hspace{-0.3 cm} \thead{\frameNb\\ \includegraphics[width=11cm, height=1.29cm]{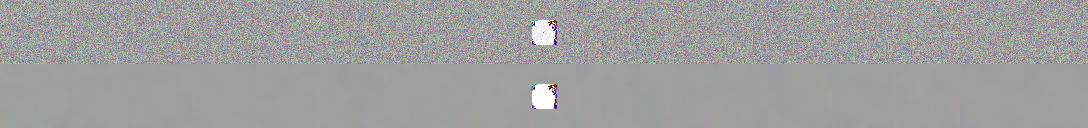}} \\
\hline
\end{tabular}
\end{table*}

\setlength{\tabcolsep}{3pt}
\begin{table*}[p]
\centering
\caption{Influence of the length of the training sequence. The table is organized in the same way as Tables~\ref{table:unconstrained_models},~\ref{table:constrained_models},~\ref{table:video-sr},~\ref{table:SRNL-3.0} and~\ref{table:feature_dampening}.}
\label{table:length_of_training_sequence}
\begin{tabular}{|c|ccc|}
\hline
\thead{7 frames} & \thead{$34.72$\\$57$\\$75$} & \thead{\includegraphics[width=3cm,height=1.7cm]{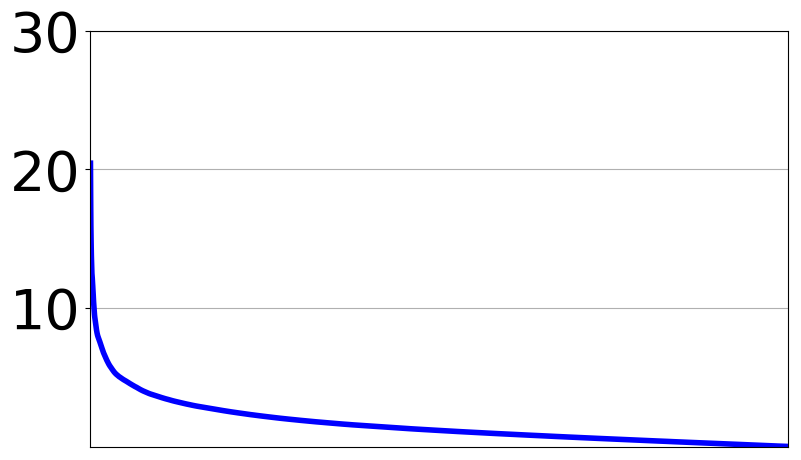}} & \thead{\vspace{-0.55cm}\\ \scriptsize $t$\vspace{0.25cm}\\ \scriptsize $X$\vspace{0.25cm}\\ \scriptsize $Y$} \hspace{-0.3 cm} \thead{\frameNb\\ \includegraphics[width=11cm, height=1.29cm]{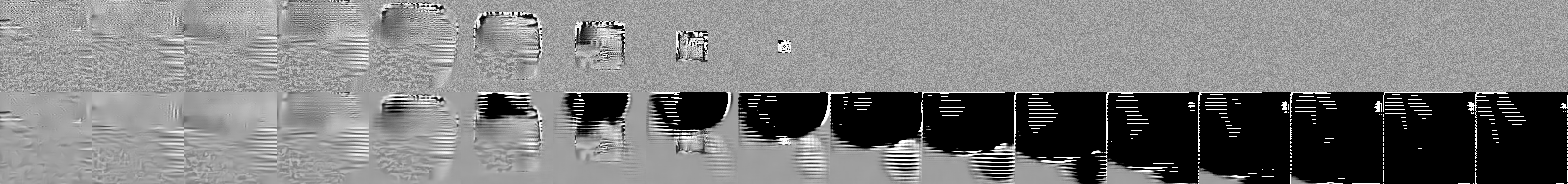}} \vspace{-0.2cm}\\
\thead{14 frames} & \thead{$34.78$\\$50$\\$7121$} & \thead{\includegraphics[width=3cm,height=1.7cm]{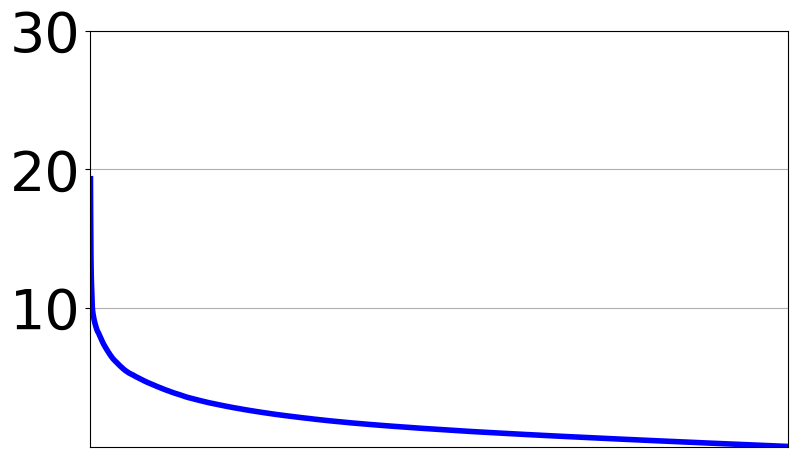}} & \thead{\vspace{-0.55cm}\\ \scriptsize $t$\vspace{0.25cm}\\ \scriptsize $X$\vspace{0.25cm}\\ \scriptsize $Y$} \hspace{-0.3 cm} \thead{\frameNb\\ \includegraphics[width=11cm, height=1.29cm]{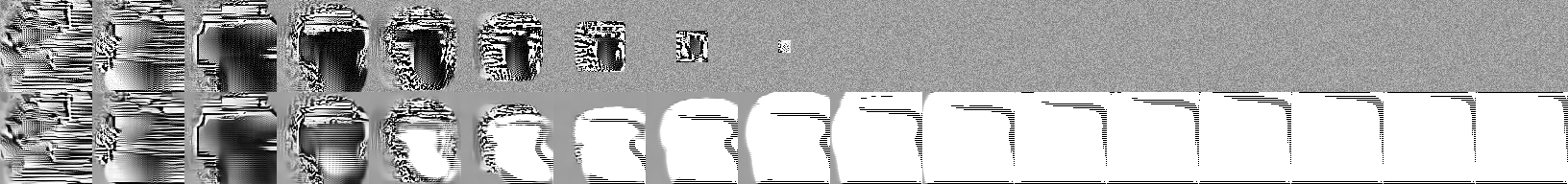}} \vspace{-0.2cm}\\
\thead{28 frames} & \thead{$34.73$\\$57$\\$234$} & \thead{\includegraphics[width=3cm,height=1.7cm]{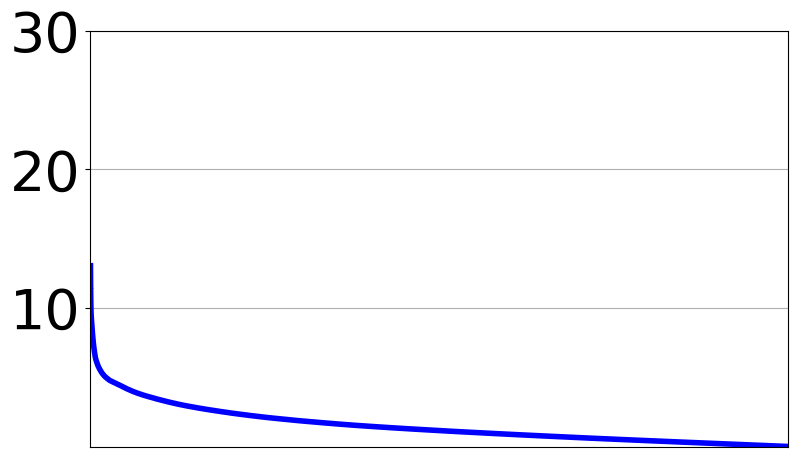}} & \thead{\vspace{-0.55cm}\\ \scriptsize $t$\vspace{0.25cm}\\ \scriptsize $X$\vspace{0.25cm}\\ \scriptsize $Y$} \hspace{-0.3 cm} \thead{\frameNb\\ \includegraphics[width=11cm, height=1.29cm]{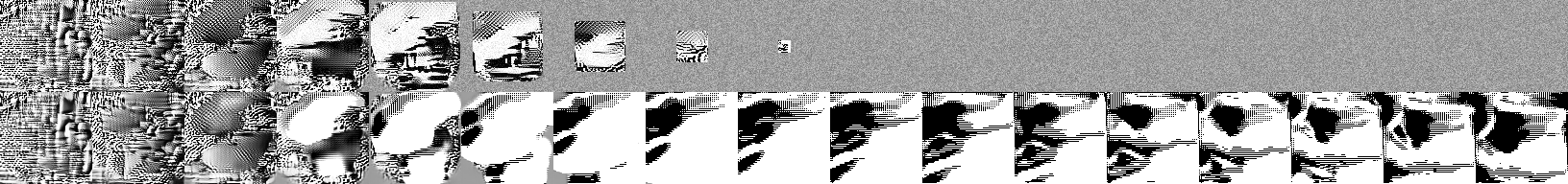}} \vspace{-0.2cm}\\
\thead{56 frames} & \thead{$34.58$\\$261$\\$12356$} & \thead{\includegraphics[width=3cm,height=1.7cm]{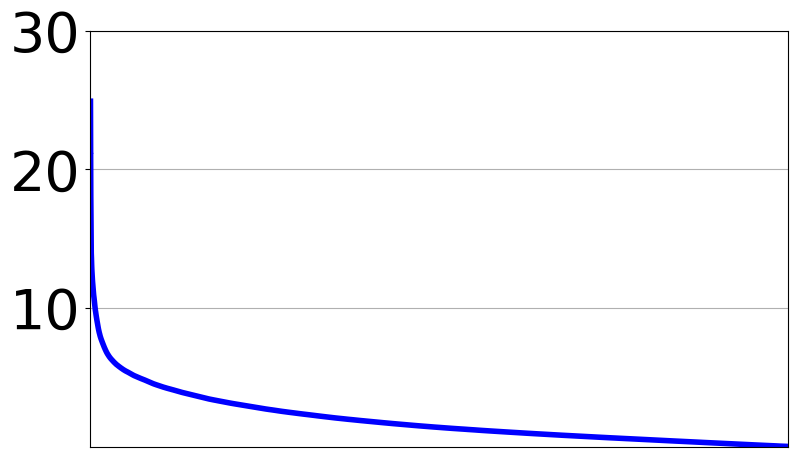}} & \thead{\vspace{-0.55cm}\\ \scriptsize $t$\vspace{0.25cm}\\ \scriptsize $X$\vspace{0.25cm}\\ \scriptsize $Y$} \hspace{-0.3 cm} \thead{\frameNb\\ \includegraphics[width=11cm, height=1.29cm]{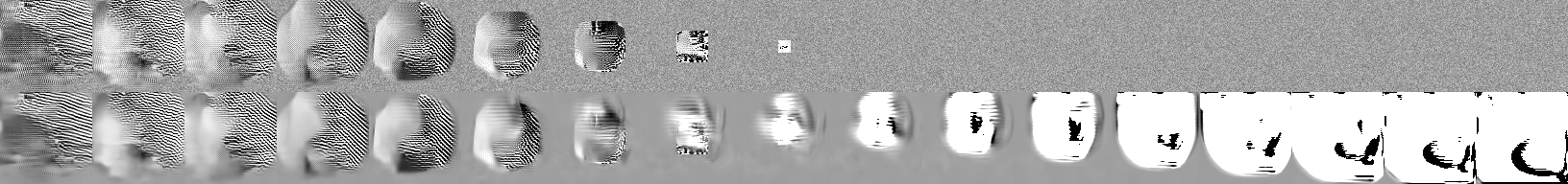}} \\
\hline
\end{tabular}
\end{table*}

\section*{Appendix G: Training on long sequences}

\begin{figure}[ht]
\begin{center}
\includegraphics[width=0.5\textwidth, height=0.4\textwidth]{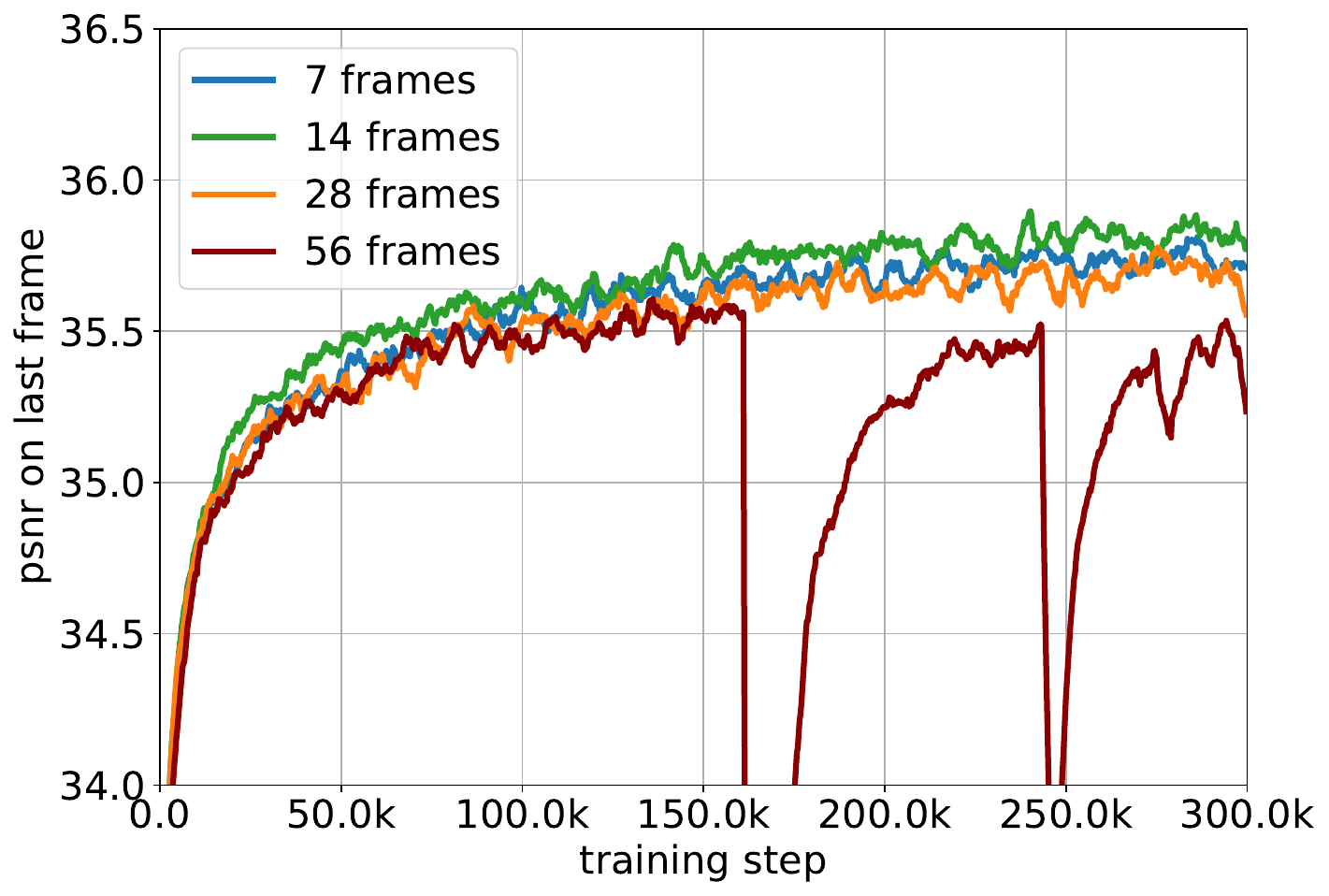}
\end{center}
\vspace{-0.4cm}
\caption{Validation PSNR on synthetic motion sequences as a function of the training step for four models trained on sequences of varying lengths.}
\label{training_length_train}
\end{figure}

In their brief discussion of the instabilities affecting their model, Godard et al. \cite{godard2018deep} suggested that they were due to an inability of the recurrent models to generalize beyond their training length sequences. To test this hypothesis however, we face computational and data constraints. It is unrealistic to train large recurrent video denoising models on sequences of more than 10 to 20 frames---the training process involves backpropagation through time, which has large memory requirements---and even if it was possible, collecting the required data would quickly become impractical. To work around these issues, we perform experiments on a small VDnCNN model where the number of internal convolutions has been reduced to only 1. This allows us to unroll the model up to 56 times through time during training. We also generate long sequences with synthetic motion from single frames in Vimeo-90k using the technique described in~\cite{boracchi2012modeling}. We train our models on gray-scale patches of $32 \times 32$ pixels using Gaussian noise with standard deviation $\sigma = 20$, for 300k training steps. We show in Figure~\ref{training_length_train} the training curves of four models trained on sequences of 7, 14, 28 and 56 frames.  Their profile is similar for the first three models but we observe sharp drops in the training curve of the model trained on sequences of 56 frames, likely due to the onset of instabilities during training resulting in gradient explosions. In Table~\ref{table:length_of_training_sequence}, we report the performance and stability of each model. Even the model trained on sequences of 56 frames is vulnerable to instabilities.

\section*{Appendix H: Influence of scene changes}

\begin{figure}[ht]
\begin{center}
\includegraphics[width=0.45\textwidth]{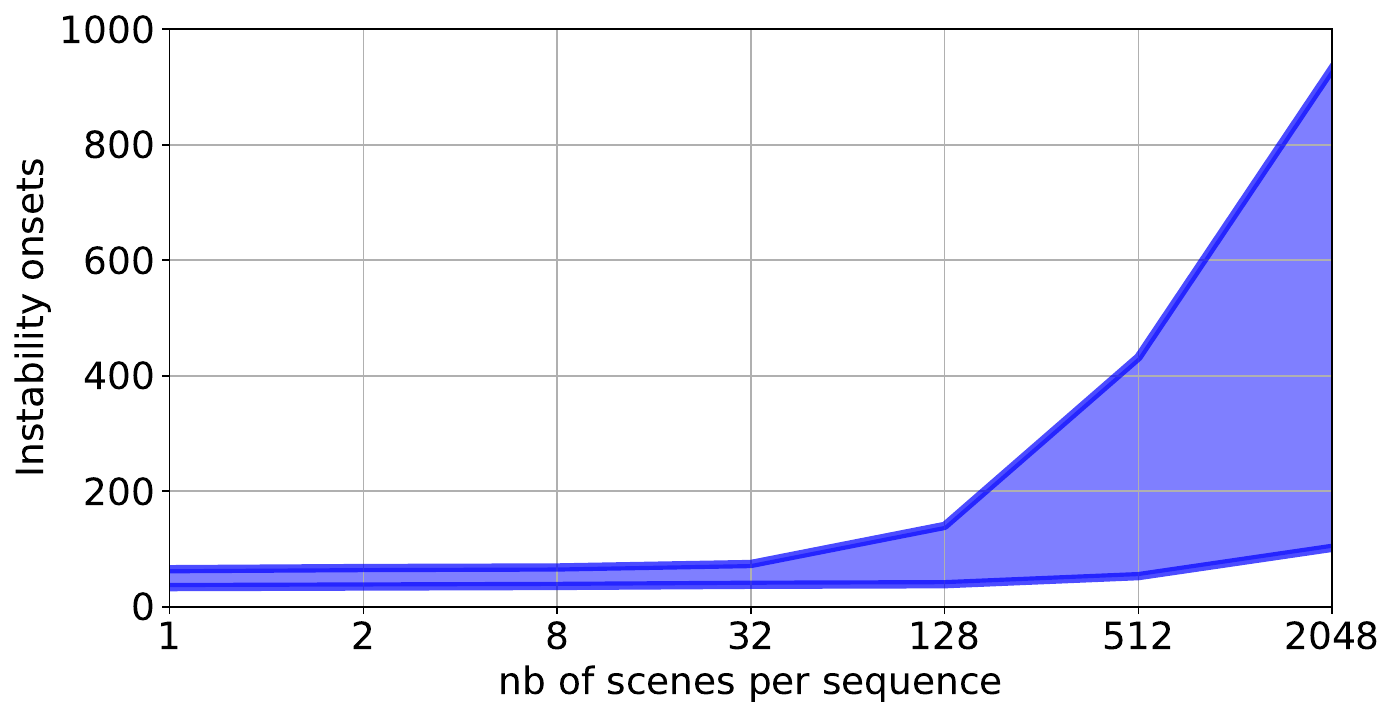}
\end{center}
\caption{$1^{st}$ and $9^{th}$ deciles of the instability onsets as a function of the number $n$ of distinct frames.}
\label{scene_changes}
\end{figure}

The long video sequences used in this paper sometimes present scene changes where the content of the video switches between two distinct scenes. Could such scene changes trigger the instabilities observed? To answer this question, we consider synthetic sequences of 2048 frames made of a number $n$ of distinct frames, randomly chosen from a large set of videos. When $n = 1$, the sequence simply consists in one long static scene. When $n=2$, the sequence presents one scene change in the middle. When $n=2048$, the sequence consists in a random succession of unrelated frames. We run VResNet-feat over such sequences a hundred times for $n \in [1, 2, 8, 32, 128, 512, 2048]$ and report the $1^{st}$ and $9^{th}$ deciles of the instability onsets in Figure~\ref{scene_changes}. Contrary to what one could expect, the instability onsets increase with the number of scene changes, i.e. VResNet-feat tends to be \emph{more stable} on sequences with scene changes. One likely explanation for this phenomenon is that scene changes interrupt the propagation of meaningful information from one frame to the next, and therefore tend to decrease the risk of positive feedback loops creating diverging outputs.

\end{document}